\documentclass[numbers,square]{article}

\usepackage{arxiv}

\usepackage[utf8]{inputenc}     % allow utf-8 input
\usepackage[T1]{fontenc}        % use 8-bit T1 fonts
\usepackage[hidelinks]{hyperref}% hyperlinks
\usepackage{url}                % simple URL typesetting
\usepackage{booktabs}           % professional-quality tables
\usepackage{amsfonts}           % blackboard math symbols
\usepackage{nicefrac}           % compact symbols for 1/2, etc.
\usepackage{microtype}          % microtypography
\usepackage{lipsum}		       % Can be removed after putting your text content
\usepackage{graphicx}
\usepackage{natbib}
\usepackage{doi}
\usepackage{xcolor}
\usepackage{dsfont}
\usepackage{float}
\usepackage{longtable}
\usepackage{array}
\usepackage[acronym]{glossaries}
\usepackage{subcaption}
\usepackage{amsmath}
\usepackage{comment}
\usepackage{caption} 
\usepackage{fancyhdr}

\fancypagestyle{firstpage}{%
  \fancyhead{}
  
  \fancyfoot[L]{Article has been accepted for publication in Propellants, Explosives, and Pyrotechnics (\href{https://doi.org/10.1002/prep.202300109}{DOI: 10.1002/prep.202300109}) \\[.1cm] DISTRIBUTION A (Log No. 23-058). Approved for Public Release; Distribution is Unlimited.}
}
\newcolumntype{L}[1]{>{\raggedright\let\newline\\\arraybackslash\hspace{0pt}}m{#1}}
\newcolumntype{C}[1]{>{\centering\let\newline\\\arraybackslash\hspace{0pt}}m{#1}}
\newcolumntype{R}[1]{>{\raggedleft\let\newline\\\arraybackslash\hspace{0pt}}m{#1}}

\DeclareMathOperator*{\argmin}{arg\,min}

\setacronymstyle{long-short}

\newacronym{ai}{AI}{Artificial Intelligence}

\newacronym{bert}{BERT}{Bidirectional Encoder Representations from Transformers}

\newacronym{cbow}{CBOW}{Continuous Bag of Words}

\newacronym{dl}{DL}{Deep Learning}

\newacronym{gpt}{GPT}{Generative Pre-trained Transformer}

\newacronym{lda}{LDA}{Latent Dirichlet Allocation}
\newacronym{llm}{LLM}{Large Language Model}
\newacronym{lstm}{LSTM}{Long Short-Term Memory}

\newacronym{ml}{ML}{Machine Learning}

\newacronym{nlp}{NLP}{Natural Language Processing}

\newacronym{sgd}{SGD}{Stochastic Gradient Descent}
\newacronym{sme}{SME}{Subject Matter Expert}

\newacronym{w2v}{W2V}{Word2Vec}

\title{Natural Language Processing for Knowledge Discovery and Information Extraction from Energetics Corpora}

\author{ \href{https://orcid.org/0000-0003-4765-6499}{\includegraphics[scale=0.06]{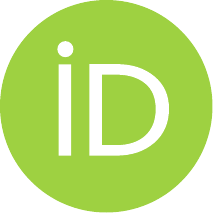}\hspace{1mm}Francis G. VanGessel}\thanks{Corresponding author (\texttt{francis.g.vangessel.civ@us.navy.mil})} , Efrem Perry, Salil Mohan, \href{https://orcid.org/0000-0002-4700-9905}{\includegraphics[scale=0.06]{orcid.PDF}\hspace{1mm}Oliver M. Barham}, Mark Cavolowsky \\
	Naval Surface Warfare Center\\
	Indian Head Division\\
	Indian Head, MD 20640
}

\renewcommand{\shorttitle}{\textit{arXiv} Template}

\hypersetup{
pdftitle={\gls{nlp} for Knowledge Discovery and Information Extraction from Energetics Corpora},
pdfsubject={},
pdfauthor={Francis VanGessel},
pdfkeywords={Energetics, Detonation Science, \gls{nlp}, Information Extraction, Knowledge Discovery},
}

\begin{document}
\maketitle
\thispagestyle{firstpage}

\fancyfoot[LE, LO]{DISTRIBUTION A (Log No. 23-058). Approved for Public Release; Distribution is Unlimited.}

\begin{abstract}
    We present a demonstration of the utility of \gls{nlp} for aiding research into energetic materials and associated systems. The \gls{nlp} method enables machine understanding of textual data, offering an automated route to knowledge discovery and information extraction from energetics text. We apply three established unsupervised \gls{nlp} models: Latent Dirichlet Allocation, Word2Vec, and the Transformer to a large curated dataset of energetics-related scientific articles. We demonstrate that each \gls{nlp} algorithm is capable of identifying energetic topics and concepts, generating a language model which aligns with Subject Matter Expert knowledge. Furthermore, we present a document classification pipeline for energetics text. Our classification pipeline achieves 59-76\% accuracy depending on the \gls{nlp} model used, with the highest performing Transformer model rivaling inter-annotator agreement metrics. The \gls{nlp} approaches studied in this work can identify concepts germane to energetics and therefore hold promise as a tool for accelerating energetics research efforts and energetics material development.
\end{abstract}

% keywords can be removed
\keywords{Energetics \and Detonation Science \and \gls{nlp} \and Knowledge Discovery \and Large Language Models}

\glsresetall
\section{Introduction}
\label{sec:introduction}
The study of energetics necessarily involves numerous scientific domains, spanning shock physics and detonation science, fluid dynamics, material science, thermodynamics, and chemical synthesis. The plethora of sub-disciplines of math, physics, chemistry, and engineering pose a challenge to practitioners who would wish to amass an expertise of energetics. Furthermore, maintaining awareness of advancements in energetics research is complicated by the exponential rate at which new research is published across scientific disciplines, including energetics. Thus, the development of automated and intelligent approaches for extracting knowledge from papers, reports, textbooks, and patents related to energetics could aid researchers and accelerate progress in energetics science.

\gls{nlp} is a sub-field of linguistics, computer science, and \gls{ml} involving the interactions between computers and human (natural) languages. \gls{nlp} techniques are used to analyze and generate human language, allowing computers to read, interpret, and understand text and speech. In the context of energetics research, \gls{nlp} can be used to analyze large volumes of textual data, such as scientific papers, technical reports, and patents, in order to extract relevant information about the concepts that underlie and explain energetics phenomenon. Furthermore, \gls{nlp} can enable natural language understanding that could be further applied to text mining journal articles and performing numerous natural language tasks such as classification, summarization, and recommendation. Overall, the use of \gls{nlp} in energetics research has the potential to enhance our understanding of energetic materials and phenomenon, and assist in the development novel propellants, explosives, and pyrotechnics.

Motivation for investigating the application of \gls{nlp} to energetics research stems from two principle sources. First, there exist a modest, but growing, promising set of \gls{ml}-based non-\gls{nlp}  studies of energetics. These studies have primarily focused on prediction of detonation properties (e.g. detonation velocity and pressure) and energetic molecule discovery \cite{elton2018applying, barnes2018machine, balakrishnan2021locally}. Despite the promising results of ML research efforts, there exists only a limited number of \gls{nlp}-based studies of energetics \cite{elton2019using,puerto2022assessing}. Second, we are motivated by the success observed in adjacent scientific fields in applying \gls{nlp} techniques to domain-specific corpora. Examples include the use of \gls{nlp} to assist in answering health related questions in the biomedical domain \cite{beltagy2019scibert} and the study of glass materials and novel thermoelectrics in the material science domain \cite{venugopal2021looking,tshitoyan2019unsupervised}. We hypothesize that \gls{nlp} approaches could provide similar benefit to research and development in the energetics domain.

In this work we asses \gls{nlp} models on their capability for extracting knowledge from, as well as classify, energetics textual data. Namely, we assess and interpret three unsupervised \gls{nlp} models on their ability to synthesize information related to the study of energetics: \gls{lda}, \gls{w2v}, and Transformers. We subsequently evaluate the downstream performance of these models as a knowledge discovery tool for classifying abstracts into energetic subdomains. We demonstrate that each \gls{nlp} model is capable of extracting critical energetics concepts and classifying energetic documents, and are therefore a useful information extraction tool.

The organizational flow of this paper proceeds as follows. Section \ref{sec:litrev} reviews the related literature, section \ref{sec:nlpmodel} provides an overview of relevant \gls{nlp} models, section \ref{sec:methods} describes the datasets, training methodologies, and evaluation procedures used in this work, section \ref{sec:results} presents and analyzes the results, and section \ref{sec:conclusion} concludes and summarize the outcomes of this research.

\section{Literature Review}
\label{sec:litrev}

In this section we provide a brief, non-exhaustive, overview of prior applications of ML and \gls{nlp} to energetics, as well as briefly survey applications of \gls{nlp} to adjacent scientific domains. It is established that training traditional ML models typically requires datasets containing hundreds to thousands of samples, while \gls{dl} models have more extensive data requirements \cite{hastie2009elements}. Requirements for ample datasets may pose a challenge when applying ML and DL methods to energetics where, because of safety and security concerns, datasets are typically small or difficult to obtain. Despite these challenges, there are a burgeoning number of studies applying ML techniques to prediction of energetic properties. Initial studies explored numerous feature representation strategies and data fusion approaches to overcome data scarcity challenges and enable training ML regression models on small datasets of 100-400 energetic molecules \cite{elton2018applying, barnes2018machine, boukouvalas2018independent, boukouvalas2021independent}. These models achieved accurate prediction of energetic properties such as density, detonation velocity, and detonation pressure. More recently, complex \gls{dl} models have been trained on thousands of molecules to predict properties such as density, detonation pressure, detonation velocity, and impact sensitivity \cite{casey2020prediction, nguyen2021predicting, lansford2022building}. These \gls{dl} models achieve high accuracy predictions, in part, due to the inclusion of energetic-like molecules that exhibit similar atomic composition and bonding motifs to known energetics. In addition to property prediction models, recent efforts have focused on the development of generative \gls{ml} techniques to predict novel energetic molecular structures \cite{balakrishnan2021locally, li2022correlated}. Generative models learn a latent representation for the probability distribution of energetic molecules, and sampling from this distribution then yields new molecules with similar, but distinct, structure to known high-explosive molecules. \gls{ml} models have also been applied to heterogeneous energetic materials, incorporating morphological information related to cracks, voids, and pores to enable the prediction of properties such as energy localization and ignition \cite{nassar2019modeling, chun2020deep}. 

Despite the varied and growing number of ML applications in the energetics domain, to the authors' knowledge there exist only two studies that apply \gls{nlp} techniques to energetics. Elton et al. utilized an \gls{nlp} algorithm to transform words appearing in energetics literature into an embedding space and demonstrated that these word embeddings could capture chemical and application relationships amongst explosive formulation material ingredients \cite{elton2019using}. Puerto et al. developed an \gls{nlp} pipeline for classifying energetic documents, utilizing three \gls{nlp} algorithms trained on a small dataset of energetic article abstracts. This work, while related to previous efforts, is broader in scope. In contrast to Elton et al. \cite{elton2019using}, we do not restrict our study of energetics solely to explosive formulation ingredients and application areas, but rather study a wider array of energetic sub-disciplines while simultaneously using more expressive \gls{nlp} models. Furthermore, we build on the research efforts of Puerto et al. \cite{puerto2022assessing} by utilizing an order of magnitude larger text corpus for training our \gls{lda} and \gls{w2v} models, explore fine-tuning of Transformer models, and perform extensive \gls{sme} assessment of the energetics information extracted by various \gls{nlp} algorithms.

While there exists a paucity of \gls{nlp} applications to energetics specifically, \gls{nlp} has shown to be a promising research tool in adjacent domains. In their review of \gls{nlp} applications to materials research, Olivetti et al. identified a common set of approaches to \gls{nlp} utilization including content acquisition, text preprocessing, document segmentation, entity recognition, and entity relation and linking \cite{olivetti2020data}. Recently, \gls{nlp} has been shown to provide a viable route to materials discovery with targeted properties. Namely, Tshitoyan et al.  performed joint word embeddings of thermoelectric and, broadly, solid-state materials; subsequent exploration of the embedding space identified candidate materials with promising thermoelectric performance \cite{tshitoyan2019unsupervised}. A notable example of \gls{nlp} applied to the scientific domain is SciBERT, a Transformer \gls{nlp} model fine-tuned on a corpus of computer science and biomedical domain papers \cite{beltagy2019scibert}. The authors demonstrate that this language model outperformed its domain-agnostic counterpart in tasks such as information extraction from clinical documents and identifying relationships between entities in computer science research articles. In a similar effort, Gupta et al. fine-tuned a Transformer-based language model on a corpus of material science publications, this model was then applied to named entity recognition, relation classification, and abstract classification in the material science research field \cite{gupta2022matscibert}. \textcolor{black}{Similar fine-tuning and entity recognition tasks were explored by Shetty et al., who applied Transformer-based models to construct polymer datasets through extraction of polymer types and associated properties such as tensile strength and molecular weight \cite{shetty2023general}. Guo et al. used Transformers to identify products and entity roles (e.g. reaction type, reactants, temperature, and yield) to create an automated reaction schema extraction pipeline \cite{guo2021automated}}. Galactica is a large language model trained on a large and broad scientific corpus including full text articles from arXiv, Scientific Scholar, and ChemRxiv, as well as other text modalities such as computer code, citations, and chemical and biological sequences \cite{taylor2022galactica}. Galactica exhibited natural language reasoning abilities across scientific domains such as math, physics, and chemistry entrance exam question answering, chemical reaction prediction, and citation generation. The success observed in \gls{nlp} applications across numerous scientific disciplines supports the need to develop natural language capabilities for energetics to aid in the acceleration of energetics research. ChatGPT \cite{ChatGPT}, a domain agnostic language model, has been shown to be capable of interpreting, and in some cases accurately answering, chemistry questions posed at the undergraduate level. Similarly, ChatGPT has been used to generate software programs for a wide-array of scientific numerical algorithms, albeit with varying levels of success \cite{kashefi2023chatgpt}.

In addition to the knowledge extraction approaches detailed above, language models have found widespread utility for chemical modeling and molecule generation. A number of researchers have applied the sequence modeling and unsupervised pretraining capabilities of \glspl{llm} to text-based molecular representations, including masked component modeling \cite{Chithrananda2020-ut, Wang2019-ap, Ross2022-ka}, autoregressive modeling \cite{Bagal2022-uo}, and autoencoder modeling \cite{Honda2019-pe}, among others.  Similar to masked language models, masked component models treat hidden elements, such as atoms, bonds, or groups, of the molecular representation and ask the model to predict the hidden value.  
\citeauthor{Wang2019-ap} developed an early approach applying masking to predicting masked components of SMILES strings in their SMILES-BERT language model.
\citeauthor{Chithrananda2020-ut} introduced ChemBERTa \cite{Chithrananda2020-ut}, a Transformer-based molecular property prediction model based on the RoBERTa architecture \cite{Liu2019-mw}. They used it to  explore the effects of various model hyperparameters and found a significant improvement with increasing pre-training dataset sizes; while the custom chemistry-focused tokenization strategy, SmilesTokenizer \cite{Schwaller2020-ek}, only offered a mild improvement over Byte-Pair Encoders; and that there was no significant difference between SMILES and SELFIES, the two popular molecular string representations.
More recently, \citeauthor{Ross2022-ka} developed MolFormer which applied linear attention and relative position embeddings to develop efficient latent molecular representations from SMILES inputs.  This model was evaluated and demonstrated good performance on property prediction, molecular similarity, and attention visualization.
In contrast to the masked component modeling, \citeauthor{Bagal2022-uo} adapted the autoregressive \gls{gpt} model \cite{Brown2020-jd} to predict the next token in a SMILES string, and demonstrated good performance on benchmarks and an ability to generate molecules with desired user parameters \cite{Bagal2022-uo}.  
\citeauthor{Honda2019-pe} developed SMILES-Transformer, which used autoencoder based pretraining that minimized the cross-entropy reconstruction loss, and demonstrated good performance on a wide variety of datasets and tasks across physical chemistry, biophysics, and physiology.

\section{ML Preliminiaries \& \gls{nlp} Model Overview}
\label{sec:nlpmodel}

\subsection{ML Preliminaries}
\label{subsec:ml_preliminaries}

In this section we briefly review fundamental techniques and concepts of \gls{ml} and \gls{nlp} that are critical to our methodology and results.

\subsubsection{ML}

\gls{ml} is a sub-field of both computer science and statistics that seeks to develop and understand algorithms that improve their performance with data.  While these algorithms have been around for decades, they have recently risen in prominence due to high-profile breakthroughs \cite{Krizhevsky2012-sn, Silver2016-oh, OpenAI2023-hh}, primarily driven by the rapid increases in computing power and the data-availability provided by the internet. 
 Ultimately, the goal of \gls{ml} is to take a finite sample of a data distribution and create a function that can make generalized predictions about the full distribution.  Formally, given input data $X \subset \mathcal{X}$, output data $Y \subset \mathcal{Y}$, and loss function $\mathcal{L}$, a \gls{ml} algorithm attempts to find a function $f : \mathcal{X} \rightarrow \mathcal{Y}$ such that

\begin{equation}
    f = \argmin_{f \in \mathcal{H}}\mathcal{L}(f|X,Y)
\end{equation}

where $\mathcal{H}$ is the set of \emph{hypothesis} functions that we are optimizing and $X$ and $Y$ are our datasets.  The identification of this function $f$ is the \emph{training} process and afterwards the function $f$ would be deployed to perform \emph{inference} on future input data sampled from $\mathcal{X}$.

Unfortunately, many algorithms learn their sample data too well, and \emph{overfit} to the training data at the expense of out-of-sample, or generalization, performance.  To combat this, it is common practice to hold out some data to evaluate the model on unseen data to estimate the performance on future samples from the underlying distribution. To ensure that the evaluation data doesn't inadvertently impact the model training and bias the model evaluation, it is important to have rigorous data handling to ensure no data leakage. To prevent data leakage, researchers must properly partition the available data into three separate sets: training, validation, and testing. The training set is used to fit the model, the validation set is used to adjust hyperparameters and prevent overfitting, and the testing set is used to evaluate the final model performance. This partitioning approach eliminates data leakage, while still generating models that generalizes well to new, unseen data.  A common training-validation-test split would be 70/10/20\%, with 70\% of the available data going toward training, 10\% of the available data going to validation, and 20\% of the available data going to testing.  Additionally, to understand the sensitivity of the model to the particular holdout sample, a $k$-fold \emph{cross validation} approach is frequently used to split the data into $k$ training-validation-test permutations.  In this method, the data is split into $k$ equal-sized subsets, and each subset will serve as the test set for a single trained model on the remaining $k-1$ subsets of training data.  The model's performance is then estimated by calculating the mean and standard deviation across all $k$ training runs to give a more generalized understanding of the model's performance.

This training/validation/test split is an example of one of many adjustable parameters that control the performance of a \gls{ml} model, also known as \emph{hyperparameters}.  Examples of hyperparameters include learning rate, regularization strength, and the number of hidden layers in a neural network. While tuning hyperparameters is essential to optimize model performance and prevent overfitting, it is difficult to predict \emph{a priori} what the effect of a hyperparameter change will have on the test performance of a model.  To find the best hyperparemeters for a given application, a hyperparameter tuning step is usually conducted which either uses a simple grid search or a more complex numerical optimization technique to find an optimal configuration.

One critical--although often implicit--hyperparameter set is the \emph{priors}, or the initial beliefs about the underlying structure of the data or model parameters.  Priors serve as a starting point for model estimation and can incorporate domain knowledge or assumptions about the data generation process. By guiding the learning process, priors can help improve model performance, particularly when the available data is limited or noisy. The appropriate choice of priors is critical, as overly strong or biased priors may hinder model generalization and lead to suboptimal results. Practically speaking, priors take the form of the type of model used, as well as the internal structure of the model.  For example, the latent dimensions of a model represent underlying variables in a model that can capture hidden patterns in the data. When little prior information is known it is typical to utilize a flat, or uniform prior, assigning equal probability to all variables or outcomes.

Another class of hyperparameters are the model optimization parameters.  This can include the type of optimizer (e.g., \gls{sgd}, Adam \cite{Kingma2014-pi}, etc.), as well as the input parameters for that optimizer.  For example, selecting the correct number of iterations through the complete dataset-- also known as \emph{epochs}--is critical to ensuring good model performance, as excessive training can cause overfitting and too little training can cause underfitting. The learning rate, another important hyperparameter, dictates the step size during the optimization process in ML models. A smaller learning rate might cause slower convergence but can potentially lead to superior solutions.

\subsubsection{\gls{nlp}}
\gls{nlp} is a branch of \gls{ml} and \gls{ai} that focuses on automated and algorithmic processing of human language. It aims to enable computers to understand, interpret, and generate human language in a way that is both meaningful and contextually appropriate. \gls{nlp} encompasses a wide range of tasks, such as sentiment analysis, machine translation, text summarization, named entity recognition, and question-answering systems. 

Early algorithms in \gls{nlp} lacked the expressive power of modern \gls{dl}-based solutions, and required significant text pre-processing prior to evaluation by the learning algorithm.  Traditional text pipelines include tokenization (segmenting text into individual words, phrases, or other meaningful units, called tokens), stopword removal (eliminating common words that do not contribute significant meaning, e.g., "and", "the", "in"), stemming (reducing words to their root form by removing affixes, e.g., "running" to "run"), and lemmatization (converting words to their base or dictionary form, e.g., "better" to "good").  These techniques help reduce noise and improve the model's ability to capture meaningful patterns.  Modern approaches--such as the Transformer family of language models--generally remove most pre-processing steps, with the exception of tokenization. Pre-processing is typically considered unnecessary for these models, as the size of the training datasets is deemed large enough to mitigate the effect of noise in the data.

In addition to pre-processing, many \gls{nlp} algorithms use the process of \emph{featurization} to convert raw data into a numerical representation that can be used as input for ML models. For text data, common featurization techniques include bag-of-words, term frequency-inverse document frequency (TF-IDF), and word embeddings. Proper featurization is essential for improving model performance and interpretability.

Lastly, in \gls{dl}-based \gls{nlp} models, the context window defines the number of surrounding words considered when analyzing a target word. A larger context window captures more contextual information, potentially improving model performance, but may also increase computational complexity.

\subsection{\gls{nlp} Model Overview}
\label{subsec:nlp_overview}

In this section we provide a brief overview of three unsupervised \gls{nlp} models used in this study: \gls{lda}; \gls{w2v}; and Transformers; and additionally the supervised classification algorithm Random Forest.

\subsubsection{Latent Dirichlet Allocation}
\label{subsubsec:lda_overview}

Developed by \citeauthor{blei2003latent}, \gls{lda} is a generative probabilistic model that has been widely used in \gls{nlp} and information retrieval \cite{blei2003latent}. It is a popular technique for topic modeling, which is the process of discovering the underlying concepts or topics that exist within a collection of documents. In \gls{lda}, each document is modeled as a probability distribution over a fixed, pre-selected, number of topics, and each topic is modeled as a distribution over words present in the corpus (Figure \ref{fig:model_over_lda}). This allows the model to learn the underlying topics that exist within a collection of documents in an unsupervised manner. Thus, once the number of topics have been specified, \gls{lda} automatically learns the words that are most associated with each topic, as well as the topics most associated with each document without the need for hand labeling. This is useful for tasks such as knowledge extraction and information retrieval, where it is often difficult to know in advance what topics will be present in a collection of documents and/or hand labeling is prohibitively time consuming. \gls{lda} has been applied to a wide range of \gls{nlp} tasks, including document classification, information retrieval, and machine translation. Overall, \gls{lda} is a powerful tool for discovering the underlying structure of a collection of documents and extracting useful concepts and ideas contained within.

\subsubsection{Word2Vec}
\label{subsubsec:lw2v_overview}

\gls{w2v} is a widely-used technique for learning dense vector representations of words, also known as word embeddings \cite{mikolov2013efficient}. These embeddings are useful for a variety of \gls{nlp} tasks, including language modeling, machine translation, and information retrieval. One of the key advantages of \gls{w2v} is that it can capture semantic relationships between words based on relative locations in the embedding space. For example, vectors for words such as "CFD," "hydrocode," and "Eulerian" will be close to one another because they are all related to computational modeling techniques. This is in contrast to traditional (sparse) techniques for representing words, which often rely on one-hot encodings that do not capture any relationships between words. There are two main flavors of \gls{w2v}: \gls{cbow} and skip-gram. \gls{cbow} predicts a target word from the context of the surrounding words, while skip-gram predicts context words given a target word  (see Figure \ref{fig:model_over_w2v} for a schematic of the \gls{cbow} \gls{w2v} variant). Skip-gram tends to perform better on small datasets, while \gls{cbow} is faster to train. Overall, \gls{w2v} is a powerful and widely-used tool for learning dense vector representations of words, which can be used directly for navigating energetic concepts and for a variety of downstream \gls{nlp} tasks.

\subsubsection{Transformer}
\label{subsubsec:Transformer_overview}

\begin{figure*}[t!]
    \centering
    \begin{subfigure}[t]{0.5\textwidth}
        \centering
        \includegraphics[trim={2cm 3cm 3cm 2cm},clip,width=9cm]{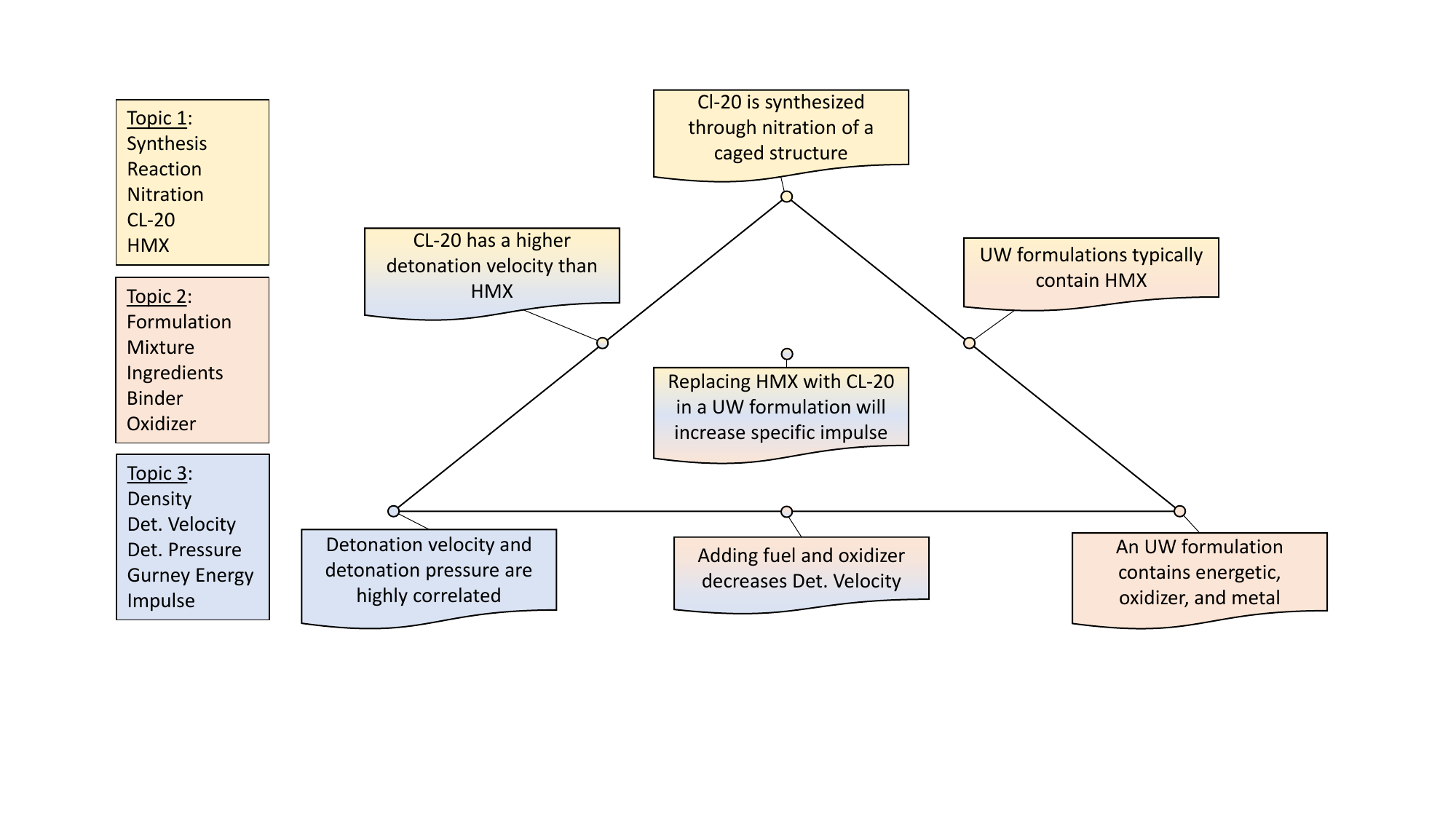}
        \caption{}
        \label{fig:model_over_lda}
    \end{subfigure}%
    ~ 
    \begin{subfigure}[t]{0.5\textwidth}
        \centering
        \includegraphics[trim={6cm 1cm 6cm 0cm},clip,width=7cm]{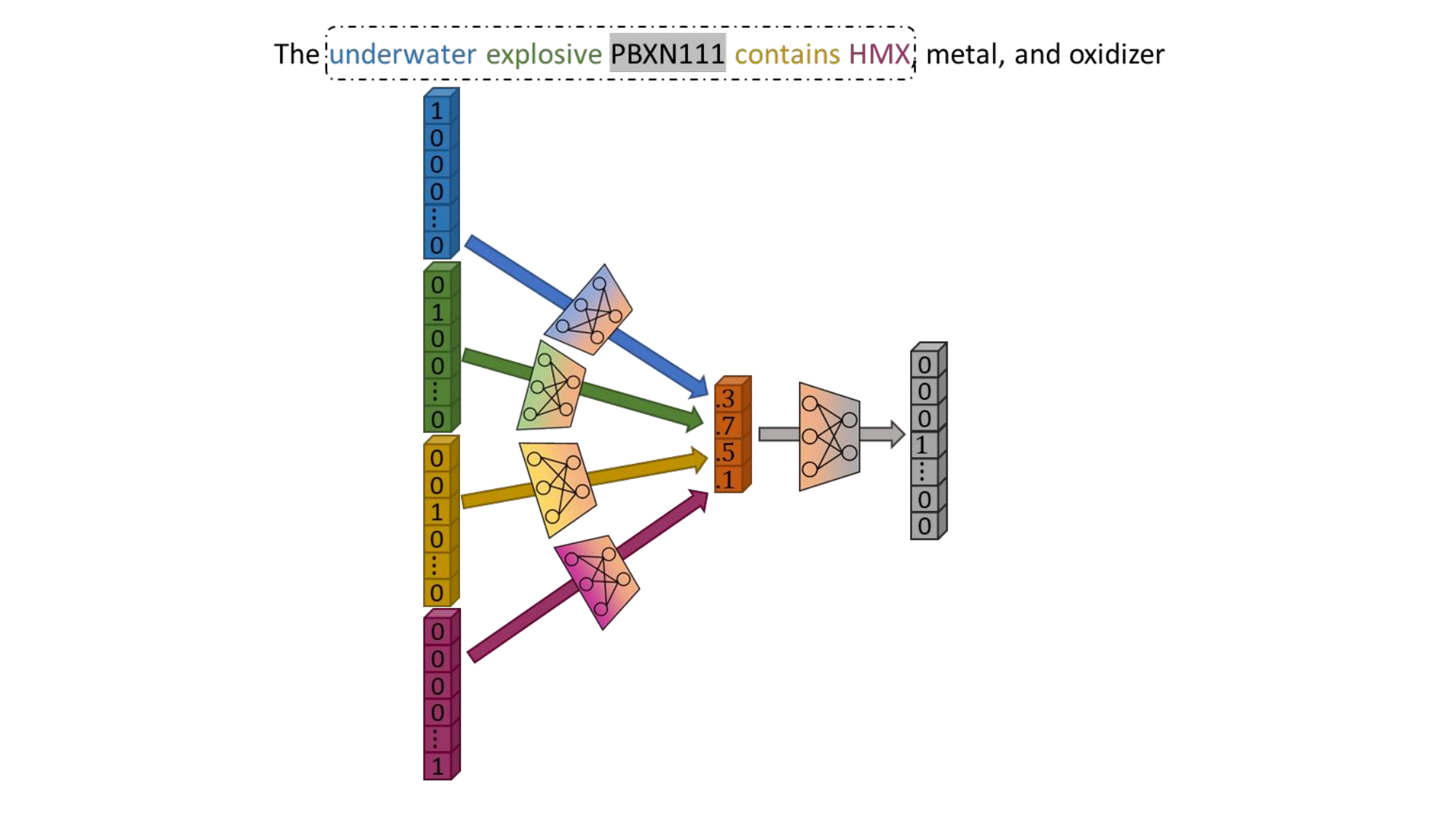}
        \caption{}
        \label{fig:model_over_w2v}
    \end{subfigure}
    \vspace{1ex}
    \begin{subfigure}[t]{0.5\textwidth}
        \centering
        \includegraphics[trim={8cm 1cm 9cm 1.5cm},clip,width=7cm]{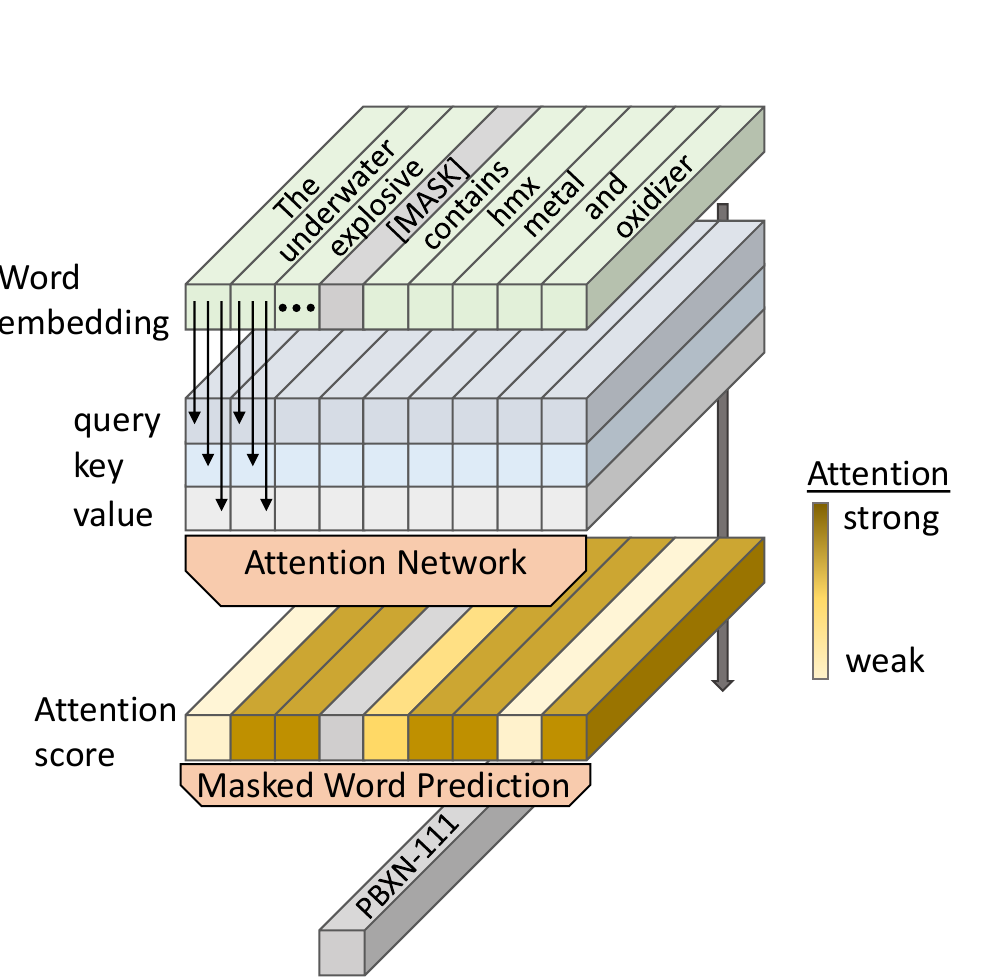}
        \caption{}
        \label{fig:model_over_Transformer} 
    \end{subfigure}
    \caption{Graphical overview of the three unsupervised \gls{nlp} models used in this study. The LDA topic model (a) assumes that each document is a mixture of topics, and each topic is represented as a probability distribution over words present in the corpus. In the example we have three topics related to molecule synthesis, energetic formulation ingredients, and explosive performance. Each document is assigned one, or more, of these topics according to the thematic elements of the document. The \gls{w2v} word embedding model (b) seeks to build a predictive model for a center word in a sequence given the surrounding context. In this example, information from every word within the context is assimilated, via individual shallow neural networks, into prediction of the center word, PBXN-111. The Transformer model (c) is trained in a self-supervised fashion to predict a masked word in a sequence. In this model, each word of a sequence is transformed into a query, key, and value vector which are combined via the \textit{attention} process into an attention score. The masked word prediction component then uses the attention score of \textit{every} word in the sequence to predicts the masked word. In contrast to the \gls{w2v} model, which equally weights each word within a finite width context window, the Transformer considers all words within the sequence, attending more strongly to informative words (e.g. explosive, HMX, and metal) while assigning less weight to uninformative words.}
\end{figure*}

Transformer language models are a class of deep neural network architectures that have been widely used for \gls{nlp} tasks such as language translation, language modeling, and text generation. Introduced in the seminal paper "Attention is All You Need" by \citeauthor{vaswani2017attention} in 2017 \cite{vaswani2017attention}, they have since become the dominant approach for these tasks. One of the key advantages of Transformer models is their ability to handle long-range dependencies in language. Traditional recurrent neural networks, such as \gls{lstm} models, can struggle to capture long-range dependencies because they process the input sequentially, one element at a time. Furthermore, \gls{w2v} uses a fixed-width context window and therefore cannot capture the long-range dependencies outside this window. In contrast, Transformer models use self-attention mechanisms to allow each element in the input to attend to all other elements, enabling them to capture long-range dependencies more effectively. Many variations of Transformer models have been proposed in the literature, including models that are designed for specific tasks such as machine translation \cite{wang2019learning} and language modeling \cite{wang2019language}, as well as models that are designed to be more efficient or to improve performance \cite{sanh2019distilbert}. Among the most common Transformer training objectives is the masked language prediction task. This approach involves randomly masking sub-word units (formally called tokens) of the corpus text and training the Transformer model to predict each hidden token (this process is in Figure \ref{fig:model_over_Transformer}). Overall, Transformer language models are a powerful and widely-used approach for a wide-array of \gls{nlp} tasks, due to their flexibility and ability to handle long-range dependencies.

\subsubsection{Random Forest}
\label{subsubsec:rf_overview}

Random Forest is a supervised \gls{ml} method commonly employed for classification and regression tasks in various domains, including \gls{nlp}. Introduced by \citeauthor{Breiman2001-kb} \cite{Breiman2001-kb}, this method creates an ensemble of decision trees to improve overall performance and mitigate the risk of overfitting. Each tree in the Random Forest is constructed independently by using a random subset of the training data and features, which introduces diversity among the individual trees. When making a prediction, each decision tree in the ensemble casts a vote, and the final decision is determined by a majority vote for classification tasks or by averaging the predictions for regression tasks. This process helps to reduce variance and improve the generalization ability of the model.  In the context of \gls{nlp}, Random Forest can be employed for tasks such as sentiment analysis, text classification, and named entity recognition. Feature extraction techniques, such as bag-of-words, TF-IDF, or word embeddings, can be used to convert text data into numerical representations suitable for the Random Forest algorithm.

\section{Methodology}
\label{sec:methods}

The methodology of this study encompasses dataset curation and preprocessing, unsupervised model training and evaluation, and supervised model training for abstract classification. Here we provide an in-depth description of each of these components.

\subsection{Data Preparation}
\label{subsec:data}

\subsubsection{Data Curation}

Our team collected text data from a range of text-based sources in an effort to develop a dataset that provides coverage of the numerous subdomains relevant to  energetics. Curation of targeted energetic domain texts is challenging due to a variety of factors including the lack of an established central repository for energetics text, online publisher security mechanisms to prevent text mining, non-standard PDF formats, and government security restrictions. Therefore, while a comprehensive final dataset has been created, it is likely non-exhaustive.

Our primary dataset consists of roughly 80,000 paragraphs and abstracts drawn from journal articles and technical reports related to energetics research. Among the sources of this data is approximately 5,000 abstracts of journal articles related to energetics, which were collated from the Journal of Energetics and the journal Propellants, Explosives, and Pyrotechnics. An additional 40,000 paragraphs, extracted from roughly 1,000 journal articles, were obtained by a custom webscraping tool developed by our team. This tool conducted a Google search using an energetics-related search term. For each website in the search results, the tool identified and extracted any associated PDF file before using additional filter terms to remove PDF files that were deemed irrelevant. We also included abstracts and paragraphs from the entire full text of the International Detonation Symposium. Finally, in order to include fundamental knowledge we scraped Wikipedia to include articles deemed relevant to energetics and explosives research. 

Our primary dataset contained a mixture of file formats, including plain text, XML and HTML, and PDF. The \gls{nlp} software libraries used in this study require plain text data as input. Thus we parsed each file using a custom file conversion pipeline that leveraged the GROBID \cite{GROBID}, Sciencebeam \cite{SCIENCEBEAM}, and BeautifulSoup \cite{BEAUTIFULSOUP} Python packages to convert the documents into plain text files. Once our data was converted to plain text format we segmented all data into individual paragraphs and abstracts for ingestion by our \gls{nlp} model training pipelines. The final dataset was randomly split into train and validation subsets for hyperparameter tuning, where the train and validation datasets contained 95\% and 5\% of the total data respectively. \textcolor{black}{Finally, the varying fidelity of document sources coupled with imperfections in the text conversion pipeline (e.g. character recognition not accurately handling older documents) resulted in instances of misspelled words in our final dataset (e.g. \emph{shook} instead of \emph{shock}). These errors effectively represent noise in our dataset and are left as is.}

A separate, hand-labeled, test dataset is also used in this study. This dataset is composed of 258 energetic materials scientific abstracts along with a class label. The classes into which the abstracts were assigned are:
\begin{enumerate}
    \item {\it Characterization}: Relating to the structural characteristics and/or performance properties of energetic materials
    \item {\it Modeling}: Relating to computational methods used to simulate the behavior of energetic materials
    \item {\it Processing}: Relating to the modifications of energetic materials or molecules
    \item {\it Synthesis}: Relating to the physical and chemical processes that produce energetic compounds
\end{enumerate}
The classes, and label assignment, were performed by an energetic materials chemistry expert. For further details on the dataset preparation procedure the reader is referred to \cite{puerto2022assessing}. This labeled dataset is used for the development of an \gls{nlp} document classification pipeline. Within this pipeline, each pretrained \gls{nlp} algorithms is used to featurize the labeled abstracts. We subsequently train a classifier to predict the abstract class label from the corresponding feature vector. The ability to rapidly classify energetic abstracts holds promise for enabling information retrieval and knowledge extraction from a large corpus.

\subsubsection{Text Preprocessing}

Each of our various \gls{nlp} algorithms, \gls{lda}, \gls{w2v}, and Transformer, require some degree of text preprocessing. We utilized the same preprocessing strategy for \gls{lda} and \gls{w2v}, this procedure consists of:
\begin{itemize}
    \item Removing newline symbols and special characters
    \item Removing stopwords and numbers
    \item Replacing variations of a chemical name with a common term
    \item Lemmatizing variations of a word into a single item
    \item Converting text to lowercase
\end{itemize}
Transformers ingest raw text by segmenting words into sub-word units, referred to as {\it tokens}, and are therefore capable of handling a more varied vocabulary than their algorithmic counterparts. Thus, we employ a minimalist approach to preprocessing the Transformer text data that includes removing newline and special characters and converting text to lowercase. 

\subsection{\gls{nlp} Model Training, Assessment, and Interpretation}
\label{sec:model_training}

\begin{figure}[ht!]
	\centering
        \includegraphics[trim={2cm 4cm 2cm 4cm},clip,width=1.0\textwidth]{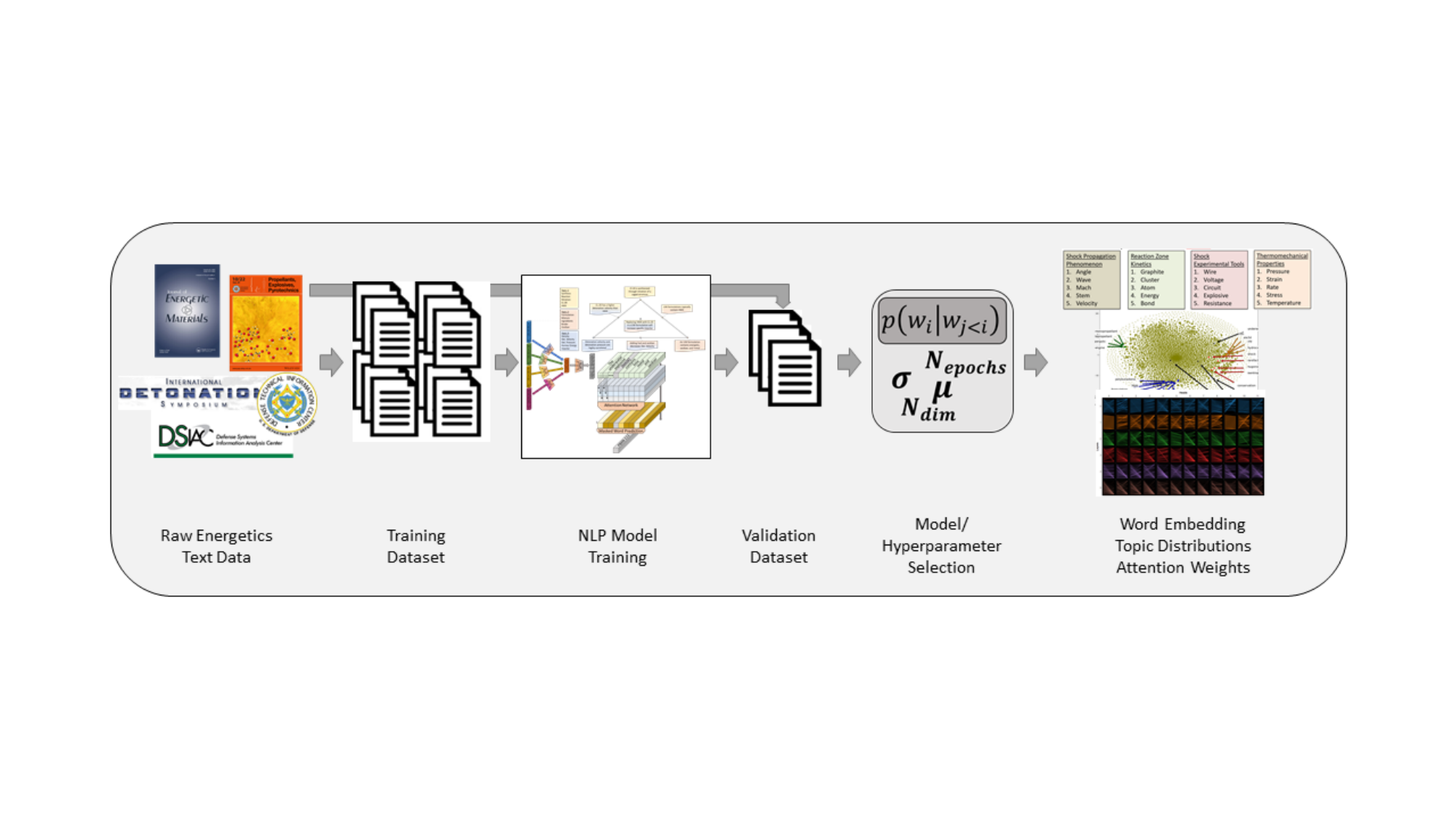}
	\caption{Training and validation procedures for \gls{lda}, \gls{w2v}, and Transformer models.}
	\label{fig:nlp_train}
\end{figure}

For each \gls{nlp} model we provide a thorough description of the training process, the hyperparameter selection and validation procedure, and protocol for assessing the models capability to extract fundamental energetics concepts. The entire workflow is depicted in Figure \ref{fig:nlp_train}.

\subsubsection{LDA}

We train all \gls{lda} models using the GENSIM Python library \cite{rehurek2011gensim}. We perform a thorough hyperparameter study including; varying the number of topics in the corpus, the prior distribution of topics within each document, and the prior distribution of words within each topic. The number of topics controls the number of latent energetic concepts present in the corpus, while the prior distributions represent the initial assumption of how topics are distributed amongst document and how words distributed amongst topics. We assess the ability of each model to generalize to new documents by measuring the average model perplexity on the held-out validation set of documents. Perplexity is a commonly used metric for evaluating how well a language model predicts a sample. As an example, consider the task of predicting the next word in the sample "their pet animal was a {\it blank}". A good language model, with low perplexity, would assign higher probabilities to words like dog, cat, or fish than to nonsensical words such as car, forest, or rain.  For \gls{lda}, we define the perplexity for a model characterized by the weights and hyperparameters, $\theta$, on the sequence of words $\mathbf{w} = (w_1, w_2, \dots, w_n)$ as:
\begin{equation}
    PPL(\mathbf{w}) = \exp \left[ -\frac{1}{n} \sum^{n}_{i=1} \log p_{\mathbf{\theta}} \left( w_i| \{w_j\}_{j<i} \right) \right] \ .
    \label{eq:perplexity}
\end{equation}
Here $n$ is the total length of the sequence and $p_{\mathbf{\theta}} \left( w_i| \{w_j\}_{j<i} \right)$ is the conditional probability the model assigns to the $i^{th}$ word given the preceding words in the sequence. Thus we can view perplexity as quantifying how well our \gls{nlp} model has learned the distribution of words present in the corpus. The hyperparameters maximizing the perplexity are given in appendix \ref{sec:appendix}.

The final \gls{lda} model is assessed on the degree to which it has absorbed concepts and ideas critical to energetics research. Specifically, an energetics \gls{sme} analyzes the ten highest probability words assigned to each topic. The the expert determines whether the words assigned to each topic represent a semantically coherent set relating to a defined energetic concept. To demonstrate the utility of topic modeling, a document, drawn for the corpus, is reviewed along with the highest probability topics assigned to the document. The topics attributed by the \gls{lda} model are critiqued by the expert as to whether they bear relevance to the themes present in the document. This expert assessment is critical to ascertain how well the model predictions align with energetics subject matter expertise and establish trust in this methodology for continued use.

\subsubsection{W2V}

The \gls{w2v} models used in this study are trained using the GENSIM topic modeling Python library \cite{rehurek2011gensim}. In the model selection process we perform a thorough hyperparameter study varying; dimension of the latent embedding, width of the context window, minimum threshold of times a word must appear, as well as consider both skipgram and \gls{cbow} model variants. The embedding dimension controls the dimensionality of the dense latent representation of words in our corpus. The context window size controls the number of neighboring words to include when making predictions regarding a center word. By setting a minimum threshold on word occurrence we effectively remove words that appear infrequently. Skipgram and \gls{cbow} refer to the two primary variants of the \gls{w2v} model as described in Section \ref{subsubsec:lw2v_overview}. An intrinsic measure of model accuracy is how well the language distribution learned by the model aligns with the distribution, i.e. sequence of words, present in the text corpus. A metric quantifying such agreement is the {\it log probability}, for an individual document containing the sequence of words $\mathbf{w} = (w_1, w_2, \dots, w_n)$ the log probability of the \gls{w2v} model, characterized by the weights and hyperparameters $\theta$, is given by the formula

\begin{equation}
   \log p_{\mathbf{\theta}}(\mathbf{w}) = \sum_{j=1}^{n} \sum_{k=1}^{n} \mathds{1}_{[1 \le |k-j| \le m ]} \log p_{\theta} (w_k|w_j) \ .
\end{equation}

Here $n$ is the document sequence length, $m$ is the context window, the indicator function $\mathds{1}$ 
is one for words falling within the context window and zero elsewhere, and $p_{\theta}$ is the learned probability distribution of the \gls{w2v} model. We select the hyperparameter combination that yields the highest median log-probability on the held out validation set of documents, these hyperparameters are given in appendix \ref{sec:appendix}.

The final \gls{w2v} model is analyzed by an energetic \gls{sme} to assess the degree to which the learned embeddings cluster semantically and conceptually related words. Words associated with a coherent energetic concept or idea should be assigned proximate embeddings in the word vector space. To rigorously assess similarity, we calculate the 10 closest embeddings for each word using the cosine similarity metric. The cosine similarity, a prevailing similarity metric in \gls{nlp}, between words $\mathbf{w_i}$ and $\mathbf{w_j}$ is

\begin{equation}
   S_C(\mathbf{w_i}, \mathbf{w_j}) = \frac{\mathbf{w_i}\cdot\mathbf{w_j}}{||\mathbf{w_i}|| ||\mathbf{w_i}||} .
   \label{eq:cos_sim}
\end{equation}
 where $||\mathbf{w}||$ is the traditional $L_2$ vector norm. Word lists of the ten closest embeddings for each word that are deemed to align with a human expert understanding of energetics are indicative of a successful embedding approach.

\subsubsection{Transformer}

The Transformer models we apply in this study were trained using the Python-based Huggingface Transformer library \cite{wolf-etal-2020-Transformers}. The Transformer library provides access to a wide variety of Transformer model variants and enables model training on GPU architectures. Due the size and complexity of Transformer models, training a model from scratch on available resources is infeasible due to data and computational requirements. Rather, we employ a {\it fine-tuning} based approach where we leverage a pretrained model, trained on a generic (i.e. domain-agnostic) corpus, and refine its weights on our domain-specific corpus. In this scenario the pretraining and fine-tuning objective were identical, for both cases the model is optimized to perform masked language prediction as described in Section \ref{subsubsec:Transformer_overview}. The cross-entropy, $H$, is a commonly used metric for multi-class classification, i.e. predicting a token from a discrete set of possibilities (i.e. token vocabulary).  Thus, minimizing the cross-entropy corresponds to ensuring the ground-truth token is assigned the maximum probability given the surrounding context, and we therefore minimize the cross-entropy during the training optimization process. The definition for cross-entropy is

\begin{equation}
   H(x,y) = \sum_{n=1}^{N} l_n  \ ; \ l_n =  - \sum_{c=1}^{C} w_{c} \log \frac{\exp(x_{n,c})}{\sum_{c=1}^{C} \exp(x_{n,i}) } y_{n,c} .
   \label{eq:cross_entropy}
\end{equation}

where $y_{n,c}$, $x_{n,c}$, and $w_n$ are the target label, unnormalized logit output, and class weight for training sample $n$, of $N$ total training samples, and token class $c$ for token vocabulary of size $C$. The final selected model is the one which yields the maximum masked token prediction accuracy on the validation dataset. We note that the token prediction accuracy metric is highly correlated, but distinct, from the cross-entropy metric. For the Transformer model only, we avoid an exhaustive hyperparameter search due to the computational cost of model training. However, we did perform a cursory investigation of the model hyperparameters including data concatenation strategies, learning rate parameters, batch size, and number of training epochs. Our final models reflects the combination of hyperparameters which were found to yield the highest prediction accuracy (the Transformer model hyperparameters are given in Appendix \ref{sec:appendix}).

The Transformer model's ability to learn the language distribution associated with energetic text is assessed in two ways. First, we contrast our fine-tuned model with other Transformer variants, including those trained on other domain-specific texts. We assess each model's ability to predict masked words within sequences drawn from the validation energetics data set. Second, an energetics \gls{sme} interprets the masked token predictions of our fine-tuned model as well as other Transformer model variants. Namely, each model is presented with a sequence taken from the energetics literature, where a token of energetics importance has been masked.  If the Transformer has been properly aligned with the energetics domain through the fine-tuning process, then the highest probability tokens produced by the model for each masked word should reflect, to some degree, the subject matter knowledge of an energetics expert.

\subsection{Abstract Classification}
\label{subsec:rf_classification}

The final ML model considered in this work is a random forest model. For this model training procedure we use the labeled dataset of 258 energetics-related abstracts previously described in Section \ref{subsec:data}. We avoid any prolonged hyperparameter tuning for this task  and train the random forest using the Scikit-learn library (the hyperparameters used in this study are reported in Appendix \ref{sec:appendix}). A five-fold cross validation procedure, as described in \ref{subsec:ml_preliminaries}, is used and we report the average of this held-out classification accuracy score, as well as the standard deviation, across all 5 folds.

Due to fundamental differences in our final \gls{nlp} models, each produces a unique feature vector for the documents in our labeled dataset. We choose to featurize each document as follows. For the \gls{lda} model, each document is represented by the topic distribution assigned by the model. For the \gls{w2v} model, we average the word embeddings of each word present in the document to obtain a single averaged embedding vector. Similarly, for the Transformer model we pass a document through the Transformer and obtain the final layer of latent embeddings from the neural network. The final layer latent embeddings are averaged across tokens to obtain an averaged token embedding vector. Therefore, each model featurizes a document as a one dimensional vector of floating point numbers, where the dimensionality varies from model to model. 

\section{Results}
\label{sec:results}

We present a thorough examination of each \gls{nlp} model with a focus on interpreting the model outputs in the context of energetics knowledge. Furthermore, we present a comparative analysis of each \gls{nlp} algorithm in their use as a featurization method for performing document classification in the energetics domain.

\subsection{LDA}

The final, best performing, \gls{lda} model achieved a perplexity score -85.31 on the validation dataset (hyperparameters are given in appendix \ref{sec:appendix}). This model, when applied to our corpus of energetic scientific texts, is found to be capable of identifying semantically coherent topics which  map to critical energetics concepts. However, there remains aspects of the LDA model that are not immediately interpretable to an energetics expert, therefore providing areas for possible model improvement. To support this observation, we present a subset of topics created by the model. These topics, their \gls{sme}-assigned theme, and associated ten highest probability words are presented in Table \ref{tab:topic_word}. \textcolor{black}{We note that while we choose to report the ten highest probability keywords consistently across topics, the distribution of probabilities associated with these top ten keywords varies from topic to topic. In some cases, the first 3 keywords contain >90\% of the cumulative probability, while in other cases this keyword probability contribution is more evenly distributed.}

\begin{table}[ht!]
	\centering
	\begin{tabular}{lp{3.0cm}p{11cm}}
		Topic No.     & Topic Theme     & Topic Keywords \\
		\midrule
		4 & Detonation Thermodynamics  & expansion, parameter, eos, adiabatic, empirical, lee, jwl, isentrope, exponential, exponent    \\[0.6cm]
		11 & Detonable Molecules & detonation, octahydro1357tetranitro1357tetrazocine, pentaerythritoltetranitrate, unreliable, hydraulicallyactuated, methoxide, blrecrystallized, recrystallizations, epc, dmdp  \\[0.9cm]
		19 & Military Roles & american, army, service, navy, scientist, engineer, lithium, french, ordnance, today \\[0.6cm]
        36 & Scientific Analysis & measured, calculated, comparison, homogeneous, examined, computed, classical, waveform, correlated, tomography \\[0.6cm]
        39 & Formulation Engineering & composition, hexahydro135trinitro135triazine, role, composite, energetic, binder, mixing, play, capturing, fundamentally \\[0.6cm]
        40 & Microstructure Morphology & mechanism, bulk, interface, prediction, modeling, porosity, impedance, inclusion, computational, localized \\[0.6cm]
        42 & SDT Modeling Theory & hot, formation, spot, mechanic, entropy, precisely, spatial, nonlinear, reproduce, crest \\[0.6cm]
        49 & Nuclear Weapon Phenomenon & nuclear, weapon, fusion, neutron, reactor, plutonium, radioactive, demand, debris, skin \\[0.6cm]
        53 & Air Shock Phenomena & wave, angle, supersonic, mach, reflected, sonic, subsonic, reproduced, overtakes, coworkers \\[0.6cm]
        162 & Propellant Deflagration & propellant, temperature, graphite, explosive, material, high, octahydro1357tetranitro1357tetrazocine, hexahydro135trinitro135triazine, boron, pressure \\[0.3cm]
        \midrule
        29 & None & one, individual, best, interval, shift, constituent, implies, rolling, overlap, shifting \\[0.6cm]
        45& None & series, strength, failed, transmitted, material, hypothesis, picture, tensile, whilst, reveals\\
		\bottomrule
	\end{tabular}
    \caption{Selected \gls{lda} Topics and associated ten highest probability keywords. The topic number is automatically generated by the model training software and has no physical interpretation. The top ten topics are highly interpetable, however the final two topics keywords lists (separated by the horizontal black line) lack coherency and cannot be readily assigned to a specific energetic subdiscipline.}
	\label{tab:topic_word}
\end{table}

Examination of the first ten topics presented in Table \ref{tab:topic_word} reveals a clear pattern of keyword grouping in a manner immediately recognizable to those with energetics knowledge. We focus on three of these topics for further discussion. First, the topic we have labeled as detonation thermodynamics, topic 4, contains a set of keywords related to the Jones-Wilkins-Lee equation of state (JWL, parameter, eos, and exponential) as well as isentropic processes. The JWL EOS is a ubiquitous equation of state used for describing the thermodynamics of detonation products as they undergo isentropic expansion in the Taylor rarefaction region behind a detonation wave. This concept is critical to understanding the governing phenomena of explosive performance capabilities of high explosive materials. Second, we have identified topic 39 as describing the material ingredients and mechanical processes related to formulation engineering. Explosive formulations typically contain a mixture of energetic materials, such as RDX (listed here as hexahydro135trinitro135triazine as a result of text preprocessing), and binder. This composition results in a material capable of releasing large amounts of energy over very short time scales with mechanical properties required for incorporation in various defense systems. The third topic, labeled air shock phenomena, relates to a set of processes linked to shock propagation resulting from an air blast. A near-ground detonation in the atmosphere produces a shockwave which impinges on the ground. The shock wave reflected from the ground, a result of the impedance mismatch between the air and ground material properties, subsequently overtakes the initial shock, forming a vertical Mach stem. The dynamical evolution of the Mach stem, including formation, height, and strength, depends on the angle at which the initial shock intersects with the ground. While many keyword lists can be readily assigned to a distinct concept as we have highlighted above, we find that certain topics are less interpretable. The final two topics listed in Table \ref{tab:topic_word} lack coherence and do not align with an energetic concept. Thus the topics and associated keywords highlighted in Table \ref{tab:topic_word} demonstrate that \gls{lda} is well suited for extracting coherent topics that capture themes critical to energetics science. 

\gls{lda} also performs the task of identifying the mixture of topics present in documents. This capability holds promise for rapid knowledge discovery by automatically grouping documents based on relevance to one another, or to distinct topics of interest. Consider the journal article excerpt \cite{gardner1965interactions} from the Fourth International Detonation Symposium presented with the corresponding two highest probability topics in Table \ref{tab:doc_topic}. This document presents analysis and interpretation of air shock, Mach reflection, critical angles. Correspondingly, the two highest probability topics identified by the \gls{lda} model correctly capture the critical concepts related to analyses of air blast generated Mach reflection. \textcolor{black}{In general, it is observed that longer documents tend to have a larger number of topics assigned to them with probabilities greater than 2\%, while shorter, abstract-length documents are typically only assigned one to two topics above the 2\% threshold. This pattern is intuitive as the longer the document, the more likely it is to relate to multiple energetic sub-disciplines. While a 2\% posterior probability for a topic may appear relatively low, it represents a six-fold increase over the uniform topic distribution prior which assigned 0.33\% probability to each topic.} We propose that one could use \gls{lda} as a knowledge discovery aid in the study of similar phenomenon by retrieving documents with similar topic distributions. This approach generalizes across topics enabling rapid information retrieval of documents relevant to, for example, energetic molecule synthesis, shock initiation modeling, or experimental characterization of explosive performance. Therefore, the ability to extract energetics concepts present in scientific texts, form coherent topics, and organize documents based on topical relevance, indicates \gls{lda} holds clear promise for energetics knowledge discovery.

\begin{table}[ht!]
	\centering
	\begin{tabular}{lp{3.5cm}p{11cm}}
		\multicolumn{3}{p{16.0cm}}{{\bf Question}: The sonic angle of critical reflection calculated with a "polytropic" equation of state is in better agreement with experiment than the limiting angle of regular reflection calculated with the same equation. Can one conclude that the critical angle is the sonic angle, rather than the limiting angle of regular reflection?} \\[0.2cm]
        \multicolumn{3}{p{16.0cm}}{{\bf Answer}: The sonic angles for both Composition B and nitromethane are calculated to be about 1 degree less than the calculated critical angles for regular reflection and therefore are in better agreement with the experimental transition angles to Mach reflection. However, in view of the rather large variation of the critical angle (as well as the sonic angle) with the several equation-of-state representations, and without an independent means for choosing which of these is most appropriate, it is impossible for us to infer to which of the three limiting angles the experimental transition angle corresponds, i.e., to the   critical angle for regular reflection, the sonic angle, or the critical angle for Mach reflection.} \\[0.2cm]
		\midrule
        53 & Air Shock Phenomena & wave, angle, supersonic, mach, reflected, sonic, subsonic, reproduced, overtakes, coworkers \\[0.3cm]
        128 & Fluid Analysis & flow understanding unique evolution reflection simulate statistical dominated classic dissipation \\
		\bottomrule
	\end{tabular}
    \caption{A document extracted from the Fourth International Detonation Symposium in which the authors analyze their air shock study \cite{gardner1965interactions}. Below the solid black line, the two highest probability topics attributed to this document are shown with the corresponding \gls{lda} topic keywords.}
	\label{tab:doc_topic}
\end{table}

\subsection{W2V}

\begin{figure}[ht!]
	\centering
	\includegraphics[trim={5cm 3cm 6cm 3cm},clip, width=0.98\textwidth]{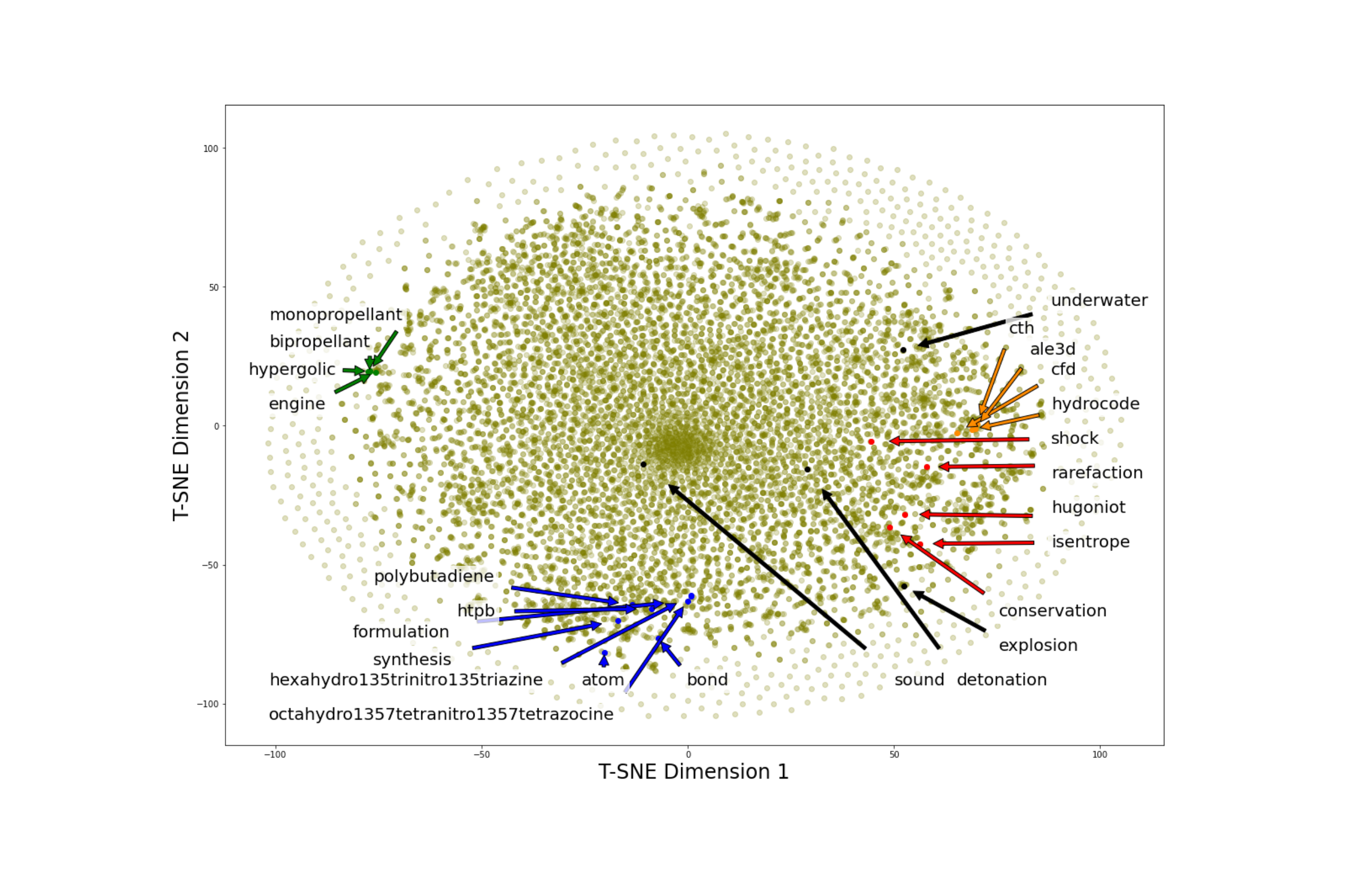}
	\caption{Word embeddings generated by \gls{w2v} model. the T-SNE algorithm had been used to reduce the dimensionality to two dimensions to aid in visualization and analysis. A select number of embeddings groups have been annotated to highlight clustering of energetic concepts. In the upper right corner, several words related to computational modeling of shock physics and high impact events have been labeled in orange. Below the modeling cluster on the right-hand side, several words related to thermodynamic processes associated with detonation dynamics have been labeled in red. At the central bottom portion of the figure, molecular synthesis concepts and high-explosive molecule names have been highlighted in blue. In the upper left corner, phrases related to propellants have been identified with green labels. Finally, certain words with no apparent thematically coherent cluster have been labeled in black.}
	\label{fig:w2v_embeddings}
\end{figure}

The final, highest performing, \gls{w2v} model achieved a median log probability on the hold out dataset of -230.74 (hyperparameters are given in appendix \ref{sec:appendix}). A thorough analysis of the model-generated word embeddings reveals that words embedded in close proximity correspond to well defined energetic motifs. To demonstrate this we visualize the word embeddings in Figure  \ref{fig:w2v_embeddings}. the T-SNE nonlinear dimensionality reduction algorithm is used to project from the 300 dimension embedding space to a two dimensional visualization space. The T-SNE algorithm operates by attempting to preserve relative distances between points when projecting from high dimensional space to the low dimensional representation. The critical tuning hyperparameter of T-SNE is the so-called perplexity (this perplexity is distinct from the \gls{nlp} variant defined in Equation \ref{eq:perplexity}) which typically ranges from 5 to 50 \cite{wattenberg2016how}, for this analysis we set perplexity to 5.

The word embeddings shown in Figure \ref{fig:w2v_embeddings} illustrate that words closely related to the same topic are embedded in close proximity. For example, the phrases "monopropellant" and "engine", labeled in green, appear close together in the left portion of the word cloud, many rocket engines utilize monopropellant based propulsion systems. In the lower center of the word cloud, labeled in blue, we see the points corresponding to the words "octahydro1357tetranitro1357tetrazocine" and "hexahydro135trinitro135triazine" nearly overlap. These are the full chemical names of the pure explosives HMX and RDX respectively, two energetics with very high detonation velocity and in widespread use in many high explosive systems. Finally, in the bottom-right quadrant of the plot, labeled in red, we observe embeddings of the words "hugoniot", "conservation", and "isentrope". These phrases refer to thermomechanical processes in which mass and momentum are conserved (hugoniot) or entropy preserving processes (isentrope). The concepts of a hugoniot and isentrope are critical to detonation theory.

While a word cloud provides a qualitative description of the \gls{w2v} generated embeddings, the dimensionality reduction can obfuscate the true nearest-neighbor embeddings of the higher dimensional space. To assess which words lie in closest proximity to one another, we refer to Table \ref{tab:w2v_cosine_similarity}. Here we have listed several words common within the field of energetics and their corresponding 10 closest word embeddings, calculated using the cosine similarity metric (Equation \ref{eq:cos_sim}). The cosine embeddings indeed reveal information loss induced by the T-SNE dimensionality reduction approach, for example "htp" a kind of propellant is placed far from "monopropellant" by the T-SNE algorithm despite monopropellant being the third most similar word. Further analysis of Table \ref{tab:w2v_cosine_similarity} reveals areas for possible model improvement as well as interesting trends. We note that many base words are embedded close to variants with differing suffix endings (e.g. bond and bonding; charge and charges; isentrope, isentopes, isentropic; etc.). Additionally, misspellings present in the original data sources pollute the embeddings: for example words like "shook" and "hock" are most likely misspelling of shock given their embedding proximity. Thus, we suspect a more thorough preprocessing of the text may further improve the quality of the \gls{w2v} generated embeddings.

Despite a few issues in the existing model, Table \ref{tab:w2v_cosine_similarity} clearly indicates that \gls{w2v} has extracted and grouped words related to critical energetic ideas. Take, for example, hydrocode which has been embedded close to names of specific software programs developed for simulation of shock physics and detonation science including; cth, ale3d, dyna3d, cale, dyna2d, and autodyn. In addition, pdv, short for photonic doppler velocimetry, is closely related, and embedded in an adjacent manner, to numerous other experimental diagnostic approaches such as visar, interferometer, and orvis. Finally, chemical bonding structures such as alkane and alkene lie close in embedding space to chemical groups commonly present in energetic materials such as carbonyl, methyl, and butyl. Thus we observe a consistent trend in which the \gls{w2v} model embeds words according to their relationship to energetic subdisciplines such as thermodynamics, explosive materials, and organic chemistry. This capability lends credence to the hypothesis that \gls{w2v}, and \gls{nlp} more generally, could enable knowledge discovery and information extraction in the energetics domain.

\begin{longtable}{p{3cm}p{12cm}}
%	\centering
%	\begin{tabular}{lp{11cm}}
%		\midrule
shock & preshock, shook, scw, shockwave, hock, steepening, overtook, ramp, precompression, reinitiation\\[0.5cm]
underwater & undex, aftermath, freefield, fae, peaceful, operability, divergent, nonideal, underground, overhaul\\[0.5cm]
isentrope & isentropes, adiabat, hugoniot, isentropic, tangency, adiabats, hnb, asymptote, hugoniots, curve\\[0.5cm]
hydrocode & cth, hydrocodes, code, ale3d, multimaterial, cfd, dyna3d, cale, dyna2d, autodyn\\[0.3cm]
conservation & continuity, conserved, thermodynamics, schrodinger, rearranging, avrami, bernoulli, governing, conserving, euler\\[0.5cm]
octahydro-1357tetranitro-1357-tetrazocine & hexahydro135trinitro135triazine, nanocrystalline, rox, phlegmatized, pentaerythritoltetranitrate, highdensity, binder, cyclotol, coarse, iimx\\[0.3cm]
propellant & propellants, motors, fueled, rocket, prilled, bipropellant, ducted, cannonball, lova, monopropellant\\[0.5cm]
deflagration & todetonation, sdt, ddt, burning, steadystate, propagation, retonation, buildup, deflagrationntoodetonation, hvd\\[0.5cm]
cfd & ale3d, numerical, coyote, amr, multimaterial, hydrocode, dyna2d, cheq, atomistic, fortran\\[0.5cm]
polybutadiene & ipdi, butadiene, ctpb, hydroxyl, nitrato, deha, terminated, pban, acrylonitrile, functionalized\\[0.5cm]
detonation & todetonation, lvd, anz, underdriven, reinitiation, backwards, hvd, superdetonation, deflagration, unsupported\\[0.5cm]
synthesis & hydrolysis, novel, itself, nitration, redox, materials, multistep, facile, particles, catalysed\\[0.5cm]
titration & styphnic, acetic, benzoic, stearic, potassiumsulfamate, citric, dinitramidic, humic, chromic, colorimetric\\[0.5cm]
cylex & ipft, floret, skid, scoping, prism, susan, subscale, freefield, steven, brazilian\\[0.5cm]
formulation & pbxs, plasticbonded, composition, cmdb, ingredient, viton, daaf, htpb, lot, aluminized\\[0.5cm]
characterization & analysis, evaluation, investigation, study, characterize, studying, assessment, characterisation, examination, fragility\\[0.5cm]
initiation & sdt, ignition, initiating, desensitisation, desensitization, ddt, coalescence, todetonation, pinch, sensitization\\[0.5cm]
bond & bonding, atom, functionalization, covalent, nitrogen, orbitals, weakest, cleavage, intramolecular, molecule\\[0.5cm]
alkane & diphenyl, ligand, alkene, moiety, dinitro, backbone, butyl, ether, carbonyl, substituted\\[0.5cm]
charge & charges, irregularly, centrally, torus, pellet, acceptor, column, egg, pear, csc\\[0.5cm]
hbx & tritonal, xtx, tnetb, btnen, pentolite, cyclotol, uncased, highdensity, tatbbased, pressed\\[0.5cm]
booster & donor, centrally, pentolite, pwb, charge, charges, ihe, acceptor, uncased, pellet\\[0.5cm]
flyer & buffer, projectile, impactor, cover, impactors, attenuator, backing, sphf, plate, driver\\[0.5cm]
engine & motor, ramjet, throttled, turbopump, ducted, powered, turbine, sounding, tripropellant, falcon\\[0.5cm]
monopropellant & bipropellant, tripropellant, monopropellants, hypergolic, candy, sounding, throttled, motors, ducted, multistage\\[0.5cm]
pdv & visar, velocimeter, ionisation, interferometer, efo, orvis, ldv, interferometry, piv, velocimetry\\[0.5cm]
	%	\bottomrule
	%\end{tabular}
    \caption{W2V model nearest neighbor embeddings calculated from cosine similarity metric.}
	\label{tab:w2v_cosine_similarity}
\end{longtable}

\subsection{Transformer}

\begin{figure}[ht!]
	\centering
	\includegraphics[trim={4cm 1cm 4cm 2cm},clip,width=0.98\textwidth]{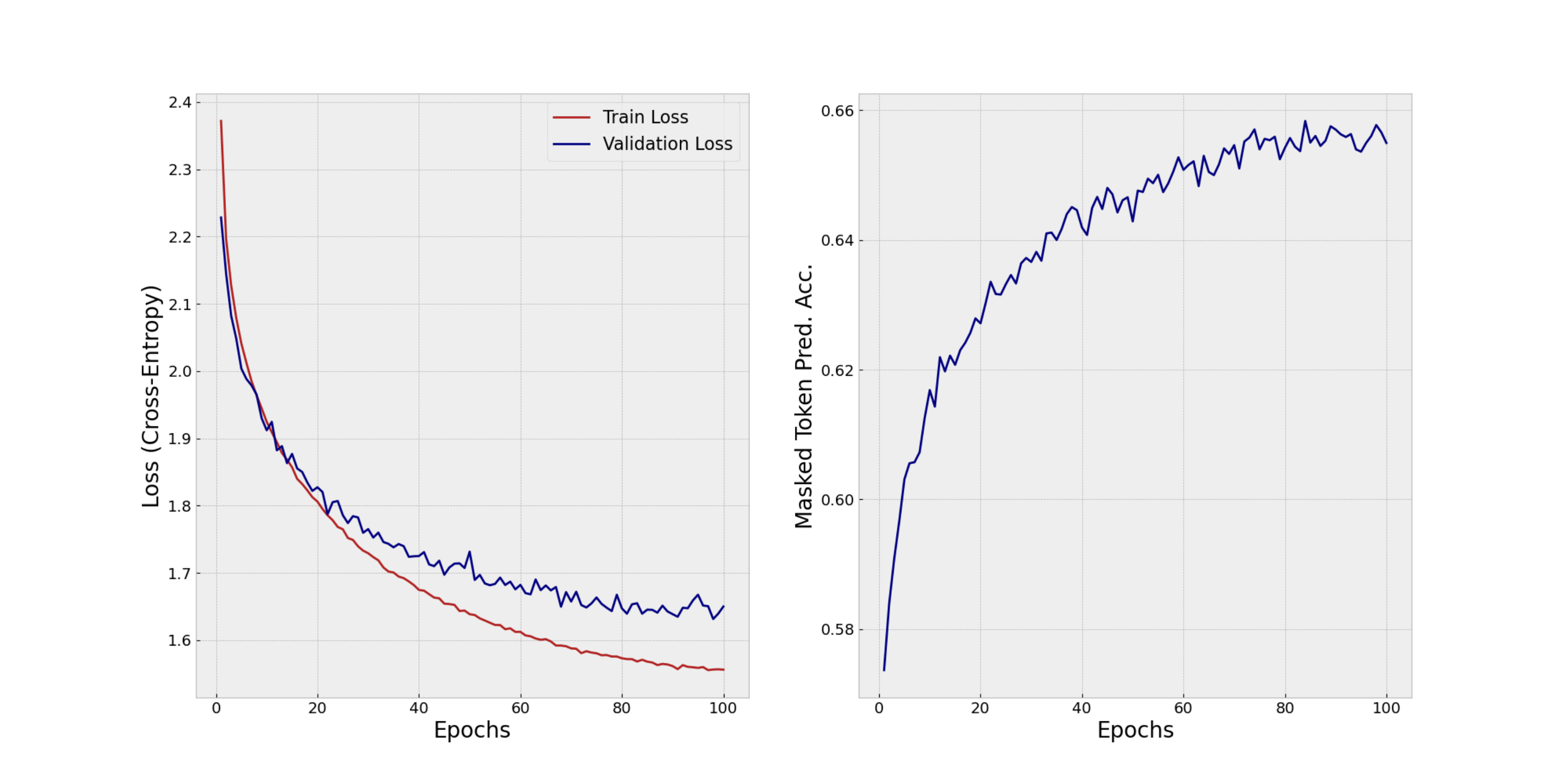}
	\caption{The cross entropy loss (calculated for training and validation data sets) and masked token prediction accuracy (calculated on only the validation data set) plotted with respect to number of epochs.}
	\label{fig:Transformer_learning_curves}
\end{figure}

Next, we attend to the Transformer \gls{nlp} model, assessing the degree to which fine-tuning the model weights can impart information critical to the energetic domain. The training procedure, including the cross-entropy loss (calculated for both the training and validation data sets) and the masked token prediction accuracy (calculated only on the validation set), is visualized in fig \ref{fig:Transformer_learning_curves}. Note that a decrease in the model loss, namely the cross-entropy, is highly correlated with an increase in the model's masked token prediction accuracy. We take the model weights from the epoch which achieved the highest masked token prediction accuracy on the validation set to be our final Transformer model, the maximum accuracy is 65.8\% which occurred at epoch 98.

To highlight the importance of fine-tuning a language model on a domain-specific corpus, in Table  \ref{tab:mlm_comparison} we contrast our energetics domain aligned model with several existing Transformer-based language models. Namely, we compare against two broad categories of models. Two of these models are domain-agnostic and trained on a massive corpus, such as the full text of English Wikipedia, these include the BERT  model and its distilled variant distilBERT \cite{sanh2019distilbert, devlin2018bert}. The remaining two models have been fine-tuned on domain-specific corpora, namely biomedicine, SciBERT, \cite{beltagy2019scibert} and material science literature related to glass materials, MatSciBERT, \cite{gupta2022matscibert}. It is apparent that the energetics fine-tuned model achieves the highest prediction accuracy. Moreover, the other scientific domain aligned models perform significantly worse in prediction accuracy, trailing even the domain-agnostic models. It would appear that despite similarities between the energetics domain and the biomedical and material science domains, there are key distinctions. These dissimilarities are made apparent by the fact that knowledge, introduced by fine-tuning a language model on a disparate domain, does not immediately transfer, leading to poor predictive performance of that model on out-of-domain tasks. This result highlights the criticality of fine-tuning a language model on a domain specific corpus when the ultimate goal is to perform natural language tasks specific to that scientific area.

\begin{table}[ht!]
	\centering
	\begin{tabular}{L{7cm}R{7cm}}
		Transformer Language Model    & Masked Token Prediction Accuracy (\%) \\
		\midrule
        Energetics Fine-tune  & \textbf{65.77}   \\[0.3cm]
		DistilBERT \cite{sanh2019distilbert}  & 58.02   \\[0.3cm]
        BERT \cite{devlin2018bert}  & 61.81   \\[0.3cm]
        SciBERT \cite{beltagy2019scibert}  & 40.78   \\[0.3cm]
        MatSciBERT \cite{gupta2022matscibert} & 40.83  \\[0.1cm]
		\bottomrule 
	\end{tabular}
    \caption{Comparison of Transformer-based language models on the masked token prediction task. The masked token prediction accuracy reported here is calculated from the validation dataset}
	\label{tab:mlm_comparison}
\end{table}

As a further comparison between domain-specific, domain-adjacent, and domain-agnostic models we investigate the predictions made for energetics-related text by the energetics fine-tuned model, SciBERT, and distilBERT. Namely, we select several sentences and mask a word of energetics relevance. These masked sequences are then passed to the three models, and each model predicts the masked word. Table \ref{tab:masked_token_prediciton_comparison} contains the results of this comparative study. The first sentence describes the Gurney equation for predicting metal acceleration capabilities of an explosive, only our fine-tuned energetics model correctly decodes the masked token. The second sentence refers to the stochiometric qualities of an explosive molecule, and both our fine-tuned model and the distilBERT model correctly identify that TNT is oxygen deficient. Finally, for our final sentence, which describes molecular structure, the energetics fine-tuned model and distilBERT are incorrect in their prediction while SciBERT correctly identifies that aromatics contain benzene rings. Based off of these results, it is clear that while opportunities for model refinement remain, the fine-tuned energetics Transformer model's attention mechanism is capable of attending to concepts germane to explosive engineering and detonation science.

\begin{table}[ht!]
	\centering
	\begin{tabular}{p{2cm}p{3cm}p{10cm}}
        Model Variant & Masked Token & Model Prediction\\
		\midrule
		Energetics Fine-tune & gurney & the {\bf gurney} method, which yields simple equations for evaluating the velocity of metals driven by detonating explosives in many geometries, is reviewed. \\[0.5cm]
        SciBERT & gurney & the {\bf analytical} method, which yields simple equations for evaluating the velocity of metals driven by detonating explosives in many geometries, is reviewed. \\[0.5cm]
        DistilBERT & gurney & the {\bf numerical} method, which yields simple equations for evaluating the velocity of metals driven by detonating explosives in many geometries, is reviewed. \\[0.3cm]
        \midrule
        Energetics Fine-tune & negative & tnt has a {\bf negative} oxygen balance. \\[0.5cm]
        SciBERT & negative & tnt has a {\bf high} oxygen balance. \\[0.3cm]
        DistilBERT & negative & tnt has a {\bf negative} oxygen balance. \\[0.1cm]
        \midrule
        Energetics Fine-tune & aromatics & molecules that contain a benzene ring are {\bf examples}. \\[0.5cm]
        SciBERT & aromatics & molecules that contain a benzene ring are {\bf aromatic}. \\[0.3cm]
        DistilBERT & aromatics & molecules that contain a benzene ring are {\bf excluded}. \\[0.1cm]
		\bottomrule
	\end{tabular}
    \caption{Masked token prediction comparison of three Transformer models. Each Transformer model is presented with the same sequence containing a masked token of energetic relevance. The resulting model prediction of that true token value is highlighted in bold.}
	\label{tab:masked_token_prediciton_comparison}
\end{table}

\subsection{Energetic Abstract Classification}

\begin{figure}[ht!]
	\centering
	\includegraphics[width=0.78\textwidth]{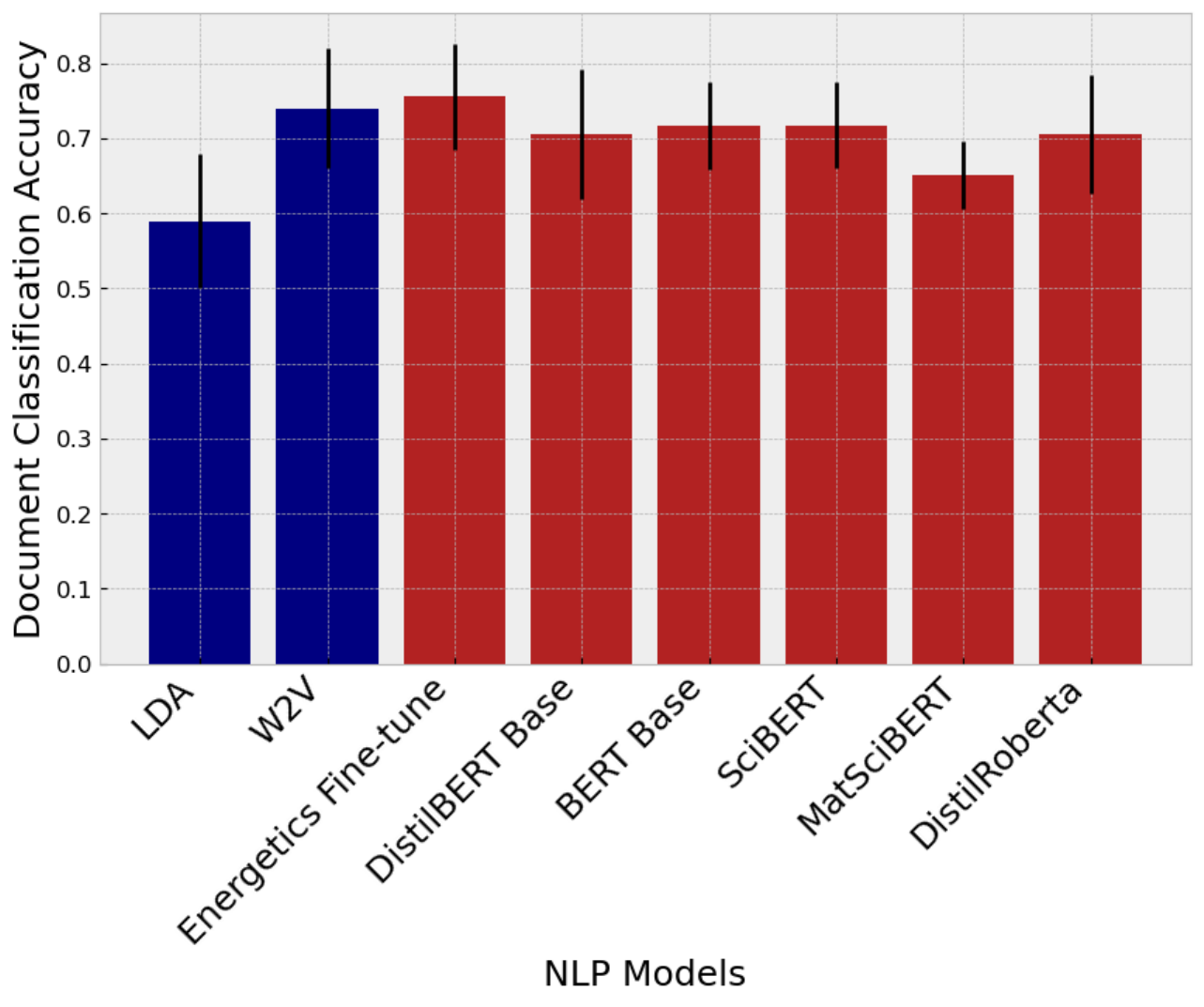}
	\caption{Document classification accuracy of energetic abstracts using various \gls{nlp} models as a featurization method. The dark red color indicates Transformer variant models. The black bar represents the standard deviation of the random forest classifier. All reported metrics are obtained from five-fold cross validation.}
	\label{fig:abstract_classification_acc}
\end{figure}

In this final section we contrast each \gls{nlp} algorithm considered thus far, namely \gls{lda}, \gls{w2v}, and the Transformer, in their capability to featurize hand-labeled abstracts pulled from the energetics literature. Namely, we use these algorithms to obtain numerical representations of the text-based abstract, this numerical feature is then used to classify each abstract into one of four distinct sub-areas of energetics via the random forest classification algorithm. Recall that the ground-truth labels of this test dataset were generated by an energetics \gls{sme} annotator. The dataset, embedding technique, and classification methodology was previously described in sections \ref{subsec:data} and \ref{subsec:rf_classification}. The accuracy of the abstract classification pipeline as well as the standard deviation of the prediction is shown in Figure \ref{fig:abstract_classification_acc}. The three leftmost entries of the plot represent the \gls{nlp} models trained in this study. The \gls{lda}, \gls{w2v}, and the Transformer model achieve 59\%, 74\%, and 76\% mean accuracy respectively, where the accuracy metrics were calculated using five-fold cross validation (the cross validation procedure was described in \ref{subsec:ml_preliminaries}). The better performance of the Transformer model is likely due to the ability of the attention mechanism to generate context-dependent embeddings for each token within an abstract, providing a more descriptive featurization and enabling the classifier to learn to  accurately assign classes. The standard deviation of the prediction accuracy ranges from 7\%-9\%, the relatively large value of the standard deviation is likely due to the small size of the test dataset which contains approximately fifty samples for each cross-validation.

\textcolor{black}{To assess the utility of developing \gls{nlp} classification approaches a naive keyword-based classifier was also developed. This keyword based approach categorizes an abstract according to which of the class label keywords, namely characterization, modeling, processing, or synthesis (or some derivative thereof), appears first in the abstract. Note that for several abstracts, none of these keywords appear and thus are left unlabeled. This keyword labeling technique classifies abstracts with 41\% accuracy, performing significantly worse than the \gls{nlp}-based approaches. It is clear that \gls{nlp} techniques provide a more informative featurization by identifying latent energetic concepts present within the text. Thus these techniques will prove useful for document retrieval, and other natural language, tasks with clear benefits over naive keyword-based retrieval techniques.}

To further understand the effect of available embedding approaches, we also assess the utility of domain-specific and domain-agnostic Transformer variants, visualized as the four right-most bars in Figure \ref{fig:abstract_classification_acc}. Our fine-tuned Transformer outperforms all other Transformer variants. However, other Transformer models do appear to provide effective embeddings, with most models achieving greater than 70\% prediction accuracy. Interestingly, the MatSciBERT variant yields the worst performance (65\% accuracy) despite being fine-tuned on an adjacent domain, i.e. materials science. Degraded performance of this model echoes the earlier observation that Transformers fine-tuned on different scientific domains perform significantly worse at the masked language prediction task (see Table \ref{tab:masked_token_prediciton_comparison}) than domain-specific or base-variant Transformers. Finally, while a classification accuracy of 100\% is in general desirable, inter-annotator agreement, even amongst experts within the scientific discipline, is not expected as abstracts could belong to multiple sub-disciplines and therefore classification thereof is subjective. The ability to rapidly ingest and classify energetic documents enables accelerated knowledge retrieval and present the possibility to assist in both research and employee training.

\section{Conclusion}
\label{sec:conclusion}

In this work we have emphasized challenges inherent in knowledge discovery and information extraction in the energetics domain. Namely, the multidisciplinary nature of energetics, which spans several scientific disciplines, and the exponential rate at which new scientific publications are published. To address these challenges, automation of knowledge discovery and information extraction is critical to enable acceleration of the pace of energetics research and energetic materials development. We propose \gls{nlp} as an automation tool for rapid analysis of energetics texts. The pursuit of an \gls{nlp}-based approach for the energetics domain was inspired by the success in adjacent domains, where \gls{nlp} has enabled text mining and expedited scientific research efforts. 

This study explored three \gls{nlp} algorithms -- \gls{lda}, \gls{w2v}, and Transformer models -- each trained on a curated corpus of over 80,000 energetic documents. The final model variants were selected after thorough hyperparameter tuning (except for the Transformer case) which involved optimizing the probabilistic language models on a held out validation set. Subsequently, an energetics \gls{sme} assessed each model's ability to incorporate concepts and ideas critical to the study of energetics. This assessment included the learned topic word distributions of \gls{lda}, the word embeddings of \gls{w2v}, and the masked language prediction of the fine-tuned Transformer model. Each model was found to incorporate knowledge gleaned from the textual data. The \gls{lda} model formed semantically cohesive topics, \gls{w2v} generated embeddings in which thematically similar words were embedded in close proximity, and the Transformer successfully predicts masked words characteristic to energetics documents. Thus, while each model exhibits aspects for further improvement, it is apparent that each yield a language model informed of concepts fundamental to the study of energetics.

Finally, to further demonstrate \gls{nlp}'s utility for knowledge retrieval and information extraction, a document classification pipeline was created. Each \gls{nlp} algorithm was used to create a document feature vector that could be used in the classification process. It was found that classification pipelines utilizing the \gls{w2v} and Transformer models achieved 74\% and 76\% classification accuracy respectively, outperforming the \gls{lda} model. Furthermore, the energetics fine-tuned Transformer achieved better performance than the other Transformer variants that were considered, highlighting the benefit of fine-tuning the network weights on a domain-specific corpus. In conclusion, this work presents a proof of concept that \gls{nlp} algorithms can serve as effective language models for the energetics discipline. The knowledge discovery and information extraction systems enabled by \gls{nlp} holds promise for accelerating the pace of energetics research and the development of new energetic materials.

\textcolor{black}{The intent of this work is to highlight the feasibility and utility of \gls{nlp} methods within energetics research. There remain nearly innumerable avenues for further exploration and development of \gls{nlp} applications. Here we highlight a few future directions we deem of critical importance. Adjacent domains such as the biomedical and material science domain have developed standardized annotated datasets containing, for example, labeled polymer entities \cite{shetty2023general} or medical question and answer pairs \cite{singhal2023large}. These datasets serve as comparative benchmarks for domain-specific natural language tasks, establishing a standardized framework for model comparison. Similarly, energetic-specific annotated datasets would support further \gls{nlp} model development within energetics. We further note that our corpus is roughly two orders of magnitude smaller in size than other \glspl{llm} training datasets \cite{beltagy2019scibert, shetty2023general}. Identifying and retrieving text data from the scattered sources characteristic of the multi-disciplinary energetics domain represents a critical advancement. Finally, we note that developments within the \gls{nlp} community over the past year have seen a rise of extremely large language models such as GPT-4 \cite{kashefi2023chatgpt} and PaLM \cite{chowdhery2022palm}. These models have three orders of magnitude more parameters than older \gls{llm}s such as BERT. For such massive language models the fine-tuning approach pursued in this work is rendered significantly more difficult due to the lack of computing power and data availability. However, the size of these models and the breadth of their datasets open alternative approaches, such as prompt tuning or few-shot learning \cite{brown2020language}, which allow these massive models to be effectively \emph{aligned} with specific domains. Aligned models achieve domain-specific comprehension, knowledge recall, and reasoning in a relatively data efficient and compute efficient manner. These alignment approaches hold promise for application of state-of-the-art language models to the energetics domain.}

\section{Author Contributions}
F.G.V conceived the project workflow, F.G.V, E.P., and S.M. collected the data, E.P. developed the webscraping tool, F.G.V and E.P developed the text preprocessing pipeline, F.G.V. developed the \gls{nlp} training and evaluation pipeline, F.G.V. analyzed the results and prepared all tables, graphs, and figures. F.G.V., E.P., and M.C. wrote the paper. O.M.B. and M.C. provided managerial oversight. All authors have reviewed, edited, and have given approval to the final manuscript.

\section*{Acknowledgement}
We thank Peter Chung (University of Maryland), Zois Boukavalas (American University), and Mark Fuge (University of Maryland) for helpful discussion. We thank Kenneth Conley as well as Will Durant, along with the Energetics Technology Center team, for facilitating the Automated Global Energetics project and providing helpful discussion. We thank Ruth Doherty and American University for generating, and sharing, labeled energetic research article abstracts. Support is gratefully acknowledged for this research from the Office of Naval Research under contract N00014-21-C-1016 and the Energetics Technology Center under contract 2054-001-002.

\section*{Data Availability Statement}
Codes and data are available upon reasonable request to
the corresponding author.

\clearpage

\bibliographystyle{unsrtnat}
\bibliography{references}  %%% Uncomment this line and comment out the ``thebibliography'' section below to use the external .bib file (using bibtex) .

\begin{thebibliography}{54}
\providecommand{\natexlab}[1]{#1}
\providecommand{\url}[1]{\texttt{#1}}
\expandafter\ifx\csname urlstyle\endcsname\relax
  \providecommand{\doi}[1]{doi: #1}\else
  \providecommand{\doi}{doi: \begingroup \urlstyle{rm}\Url}\fi

\bibitem[Elton et~al.(2018)Elton, Boukouvalas, Butrico, Fuge, and
  Chung]{elton2018applying}
Daniel~C Elton, Zois Boukouvalas, Mark~S Butrico, Mark~D Fuge, and Peter~W
  Chung.
\newblock Applying machine learning techniques to predict the properties of
  energetic materials.
\newblock \emph{Scientific reports}, 8\penalty0 (1):\penalty0 1--12, 2018.

\bibitem[Barnes et~al.(2018)Barnes, Elton, Boukouvalas, Taylor, Mattson, Fuge,
  and Chung]{barnes2018machine}
Brian~C Barnes, Daniel~C Elton, Zois Boukouvalas, DeCarlos~E Taylor, William~D
  Mattson, Mark~D Fuge, and Peter~W Chung.
\newblock Machine learning of energetic material properties.
\newblock \emph{arXiv preprint arXiv:1807.06156}, 2018.

\bibitem[Balakrishnan et~al.(2021)Balakrishnan, VanGessel, Boukouvalas, Barnes,
  Fuge, and Chung]{balakrishnan2021locally}
Sangeeth Balakrishnan, Francis~G VanGessel, Zois Boukouvalas, Brian~C Barnes,
  Mark~D Fuge, and Peter~W Chung.
\newblock Locally optimizable joint embedding framework to design nitrogen-rich
  molecules that are similar but improved.
\newblock \emph{Molecular Informatics}, 40\penalty0 (7):\penalty0 2100011,
  2021.

\bibitem[Elton et~al.(2019)Elton, Turakhia, Reddy, Boukouvalas, Fuge, Doherty,
  and Chung]{elton2019using}
Daniel~C Elton, Dhruv Turakhia, Nischal Reddy, Zois Boukouvalas, Mark~D Fuge,
  Ruth~M Doherty, and Peter~W Chung.
\newblock Using natural language processing techniques to extract information
  on the properties and functionalities of energetic materials from large text
  corpora.
\newblock \emph{arXiv preprint arXiv:1903.00415}, 2019.

\bibitem[Puerto et~al.(2022)Puerto, Kellett, Nikopoulou, Fuge, Doherty, Chung,
  and Boukouvalas]{puerto2022assessing}
Monica Puerto, Mason Kellett, Rodanthi Nikopoulou, Mark~D Fuge, Ruth Doherty,
  Peter~W Chung, and Zois Boukouvalas.
\newblock Assessing the trade-off between prediction accuracy and
  interpretability for topic modeling on energetic materials corpora.
\newblock \emph{arXiv preprint arXiv:2206.00773}, 2022.

\bibitem[Beltagy et~al.(2019)Beltagy, Lo, and Cohan]{beltagy2019scibert}
Iz~Beltagy, Kyle Lo, and Arman Cohan.
\newblock Scibert: A pretrained language model for scientific text.
\newblock \emph{arXiv preprint arXiv:1903.10676}, 2019.

\bibitem[Venugopal et~al.(2021)Venugopal, Sahoo, Zaki, Agarwal, Gosvami, and
  Krishnan]{venugopal2021looking}
Vineeth Venugopal, Sourav Sahoo, Mohd Zaki, Manish Agarwal, Nitya~Nand Gosvami,
  and NM~Anoop Krishnan.
\newblock Looking through glass: Knowledge discovery from materials science
  literature using natural language processing.
\newblock \emph{Patterns}, 2\penalty0 (7):\penalty0 100290, 2021.

\bibitem[Tshitoyan et~al.(2019)Tshitoyan, Dagdelen, Weston, Dunn, Rong,
  Kononova, Persson, Ceder, and Jain]{tshitoyan2019unsupervised}
Vahe Tshitoyan, John Dagdelen, Leigh Weston, Alexander Dunn, Ziqin Rong, Olga
  Kononova, Kristin~A Persson, Gerbrand Ceder, and Anubhav Jain.
\newblock Unsupervised word embeddings capture latent knowledge from materials
  science literature.
\newblock \emph{Nature}, 571\penalty0 (7763):\penalty0 95--98, 2019.

\bibitem[Hastie et~al.(2009)Hastie, Tibshirani, Friedman, and
  Friedman]{hastie2009elements}
Trevor Hastie, Robert Tibshirani, Jerome~H Friedman, and Jerome~H Friedman.
\newblock \emph{The elements of statistical learning: data mining, inference,
  and prediction}, volume~2.
\newblock Springer, 2009.

\bibitem[Boukouvalas et~al.(2018)Boukouvalas, Elton, Chung, and
  Fuge]{boukouvalas2018independent}
Zois Boukouvalas, Daniel~C Elton, Peter~W Chung, and Mark~D Fuge.
\newblock Independent vector analysis for data fusion prior to molecular
  property prediction with machine learning.
\newblock \emph{arXiv preprint arXiv:1811.00628}, 2018.

\bibitem[Boukouvalas et~al.(2021)Boukouvalas, Puerto, Elton, Chung, and
  Fuge]{boukouvalas2021independent}
Zois Boukouvalas, Monica Puerto, Daniel~C Elton, Peter~W Chung, and Mark~D
  Fuge.
\newblock Independent vector analysis for molecular data fusion: Application to
  property prediction and knowledge discovery of energetic materials.
\newblock In \emph{2020 28th European Signal Processing Conference (EUSIPCO)},
  pages 1030--1034. IEEE, 2021.

\bibitem[Casey et~al.(2020)Casey, Son, Bilionis, and
  Barnes]{casey2020prediction}
Alex~D Casey, Steven~F Son, Ilias Bilionis, and Brian~C Barnes.
\newblock Prediction of energetic material properties from electronic structure
  using 3d convolutional neural networks.
\newblock \emph{Journal of Chemical Information and Modeling}, 60\penalty0
  (10):\penalty0 4457--4473, 2020.

\bibitem[Nguyen et~al.(2021)Nguyen, Loveland, Kim, Karande, Hiszpanski, and
  Han]{nguyen2021predicting}
Phan Nguyen, Donald Loveland, Joanne~T Kim, Piyush Karande, Anna~M Hiszpanski,
  and T~Yong-Jin Han.
\newblock Predicting energetics materials’ crystalline density from chemical
  structure by machine learning.
\newblock \emph{Journal of Chemical Information and Modeling}, 61\penalty0
  (5):\penalty0 2147--2158, 2021.

\bibitem[Lansford et~al.(2022)Lansford, Barnes, Rice, and
  Jensen]{lansford2022building}
Joshua~L Lansford, Brian~C Barnes, Betsy~M Rice, and Klavs~F Jensen.
\newblock Building chemical property models for energetic materials from small
  datasets using a transfer learning approach.
\newblock \emph{Journal of Chemical Information and Modeling}, 62\penalty0
  (22):\penalty0 5397--5410, 2022.

\bibitem[Li et~al.(2022)Li, Wang, Sun, Zeng, Yuan, Gou, Wang, Guo, and
  Pu]{li2022correlated}
Chuan Li, Chenghui Wang, Ming Sun, Yan Zeng, Yuan Yuan, Qiaolin Gou, Guangchuan
  Wang, Yanzhi Guo, and Xuemei Pu.
\newblock Correlated rnn framework to quickly generate molecules with desired
  properties for energetic materials in the low data regime.
\newblock \emph{Journal of Chemical Information and Modeling}, 62\penalty0
  (20):\penalty0 4873--4887, 2022.

\bibitem[Nassar et~al.(2019)Nassar, Rai, Sen, and
  Udaykumar]{nassar2019modeling}
Anas Nassar, Nirmal~K Rai, Oishik Sen, and HS~Udaykumar.
\newblock Modeling mesoscale energy localization in shocked hmx, part i:
  machine-learned surrogate models for the effects of loading and void sizes.
\newblock \emph{Shock Waves}, 29\penalty0 (4):\penalty0 537--558, 2019.

\bibitem[Chun et~al.(2020)Chun, Roy, Nguyen, Choi, Udaykumar, and
  Baek]{chun2020deep}
Sehyun Chun, Sidhartha Roy, Yen~Thi Nguyen, Joseph~B Choi, HS~Udaykumar, and
  Stephen~S Baek.
\newblock Deep learning for synthetic microstructure generation in a
  materials-by-design framework for heterogeneous energetic materials.
\newblock \emph{Scientific reports}, 10\penalty0 (1):\penalty0 1--15, 2020.

\bibitem[Olivetti et~al.(2020)Olivetti, Cole, Kim, Kononova, Ceder, Han, and
  Hiszpanski]{olivetti2020data}
Elsa~A Olivetti, Jacqueline~M Cole, Edward Kim, Olga Kononova, Gerbrand Ceder,
  Thomas Yong-Jin Han, and Anna~M Hiszpanski.
\newblock Data-driven materials research enabled by natural language processing
  and information extraction.
\newblock \emph{Applied Physics Reviews}, 7\penalty0 (4):\penalty0 041317,
  2020.

\bibitem[Gupta et~al.(2022)Gupta, Zaki, Krishnan, et~al.]{gupta2022matscibert}
Tanishq Gupta, Mohd Zaki, NM~Krishnan, et~al.
\newblock Matscibert: A materials domain language model for text mining and
  information extraction.
\newblock \emph{npj Computational Materials}, 8\penalty0 (1):\penalty0 1--11,
  2022.

\bibitem[Shetty et~al.(2023)Shetty, Rajan, Kuenneth, Gupta, Panchumarti, Holm,
  Zhang, and Ramprasad]{shetty2023general}
Pranav Shetty, Arunkumar~Chitteth Rajan, Chris Kuenneth, Sonakshi Gupta,
  Lakshmi~Prerana Panchumarti, Lauren Holm, Chao Zhang, and Rampi Ramprasad.
\newblock A general-purpose material property data extraction pipeline from
  large polymer corpora using natural language processing.
\newblock \emph{npj Computational Materials}, 9\penalty0 (1):\penalty0 52,
  2023.

\bibitem[Guo et~al.(2021)Guo, Ibanez-Lopez, Gao, Quach, Coley, Jensen, and
  Barzilay]{guo2021automated}
Jiang Guo, A~Santiago Ibanez-Lopez, Hanyu Gao, Victor Quach, Connor~W Coley,
  Klavs~F Jensen, and Regina Barzilay.
\newblock Automated chemical reaction extraction from scientific literature.
\newblock \emph{Journal of chemical information and modeling}, 62\penalty0
  (9):\penalty0 2035--2045, 2021.

\bibitem[Taylor et~al.(2022)Taylor, Kardas, Cucurull, Scialom, Hartshorn,
  Saravia, Poulton, Kerkez, and Stojnic]{taylor2022galactica}
Ross Taylor, Marcin Kardas, Guillem Cucurull, Thomas Scialom, Anthony
  Hartshorn, Elvis Saravia, Andrew Poulton, Viktor Kerkez, and Robert Stojnic.
\newblock Galactica: A large language model for science.
\newblock \emph{arXiv preprint arXiv:2211.09085}, 2022.

\bibitem[OpenAI(2022)]{ChatGPT}
OpenAI.
\newblock Chatgpt: Optimizing language models for dialogue, 2022.
\newblock URL \url{https://openai.com/blog/chatgpt/}.

\bibitem[Kashefi and Mukerji(2023)]{kashefi2023chatgpt}
Ali Kashefi and Tapan Mukerji.
\newblock Chatgpt for programming numerical methods.
\newblock \emph{arXiv preprint arXiv:2303.12093}, 2023.

\bibitem[Chithrananda et~al.(2020)Chithrananda, Grand, and
  Ramsundar]{Chithrananda2020-ut}
Seyone Chithrananda, Gabriel Grand, and Bharath Ramsundar.
\newblock {ChemBERTa}: {Large-Scale} {Self-Supervised} pretraining for
  molecular property prediction.
\newblock October 2020.
\newblock URL \url{http://arxiv.org/abs/2010.09885}.

\bibitem[Wang et~al.(2019{\natexlab{a}})Wang, Guo, Wang, Sun, and
  Huang]{Wang2019-ap}
Sheng Wang, Yuzhi Guo, Yuhong Wang, Hongmao Sun, and Junzhou Huang.
\newblock {SMILES-BERT}: Large scale unsupervised {Pre-Training} for molecular
  property prediction.
\newblock In \emph{Proceedings of the 10th {ACM} International Conference on
  Bioinformatics, Computational Biology and Health Informatics}, BCB '19, pages
  429--436, New York, NY, USA, September 2019{\natexlab{a}}. Association for
  Computing Machinery.
\newblock ISBN 9781450366663.
\newblock \doi{10.1145/3307339.3342186}.
\newblock URL \url{https://doi.org/10.1145/3307339.3342186}.

\bibitem[Ross et~al.(2022)Ross, Belgodere, Chenthamarakshan, Padhi, Mroueh, and
  Das]{Ross2022-ka}
Jerret Ross, Brian Belgodere, Vijil Chenthamarakshan, Inkit Padhi, Youssef
  Mroueh, and Payel Das.
\newblock Molformer: Large scale chemical language representations capture
  molecular structure and properties.
\newblock May 2022.
\newblock URL
  \url{https://www.researchsquare.com/article/rs-1570270/latest.pdf}.

\bibitem[Bagal et~al.(2022)Bagal, Aggarwal, Vinod, and
  Priyakumar]{Bagal2022-uo}
Viraj Bagal, Rishal Aggarwal, P~K Vinod, and U~Deva Priyakumar.
\newblock {MolGPT}: Molecular generation using a {Transformer-Decoder} model.
\newblock \emph{J. Chem. Inf. Model.}, 62\penalty0 (9):\penalty0 2064--2076,
  May 2022.
\newblock ISSN 1549-9596, 1549-960X.
\newblock \doi{10.1021/acs.jcim.1c00600}.
\newblock URL \url{http://dx.doi.org/10.1021/acs.jcim.1c00600}.

\bibitem[Honda et~al.(2019)Honda, Shi, and Ueda]{Honda2019-pe}
Shion Honda, Shoi Shi, and Hiroki~R Ueda.
\newblock {SMILES} transformer: Pre-trained molecular fingerprint for low data
  drug discovery.
\newblock November 2019.
\newblock URL \url{http://arxiv.org/abs/1911.04738}.

\bibitem[Liu et~al.(2019)Liu, Ott, Goyal, Du, Joshi, Chen, Levy, Lewis,
  Zettlemoyer, and Stoyanov]{Liu2019-mw}
Yinhan Liu, Myle Ott, Naman Goyal, Jingfei Du, Mandar Joshi, Danqi Chen, Omer
  Levy, Mike Lewis, Luke Zettlemoyer, and Veselin Stoyanov.
\newblock {RoBERTa}: A robustly optimized {BERT} pretraining approach.
\newblock July 2019.
\newblock URL \url{http://arxiv.org/abs/1907.11692}.

\bibitem[Schwaller et~al.(2020)Schwaller, Probst, Vaucher, Nair, Kreutter,
  Laino, and Reymond]{Schwaller2020-ek}
Philippe Schwaller, Daniel Probst, Alain~C Vaucher, Vishnu~H Nair, David
  Kreutter, Teodoro Laino, and Jean-Louis Reymond.
\newblock Mapping the space of chemical reactions using {Attention-Based}
  neural networks.
\newblock \emph{ChemRxiv}, December 2020.
\newblock \doi{10.26434/chemrxiv.9897365.v4}.
\newblock URL
  \url{https://chemrxiv.org/engage/chemrxiv/article-details/60c753a0bdbb89acf8a3a4b5}.

\bibitem[Brown et~al.(2020{\natexlab{a}})Brown, Mann, Ryder, Subbiah, Kaplan,
  Dhariwal, Neelakantan, Shyam, Sastry, Askell, Agarwal, Herbert-Voss, Krueger,
  Henighan, Child, Ramesh, Ziegler, Wu, Winter, Hesse, Chen, Sigler, Litwin,
  Gray, Chess, Clark, Berner, McCandlish, Radford, Sutskever, and
  Amodei]{Brown2020-jd}
Tom~B Brown, Benjamin Mann, Nick Ryder, Melanie Subbiah, Jared Kaplan, Prafulla
  Dhariwal, Arvind Neelakantan, Pranav Shyam, Girish Sastry, Amanda Askell,
  Sandhini Agarwal, Ariel Herbert-Voss, Gretchen Krueger, Tom Henighan, Rewon
  Child, Aditya Ramesh, Daniel~M Ziegler, Jeffrey Wu, Clemens Winter,
  Christopher Hesse, Mark Chen, Eric Sigler, Mateusz Litwin, Scott Gray,
  Benjamin Chess, Jack Clark, Christopher Berner, Sam McCandlish, Alec Radford,
  Ilya Sutskever, and Dario Amodei.
\newblock Language models are {Few-Shot} learners.
\newblock May 2020{\natexlab{a}}.
\newblock URL \url{http://arxiv.org/abs/2005.14165}.

\bibitem[Krizhevsky et~al.(2012)Krizhevsky, Sutskever, and
  Hinton]{Krizhevsky2012-sn}
Alex Krizhevsky, Ilya Sutskever, and Geoffrey~E Hinton.
\newblock {ImageNet} classification with deep convolutional neural networks.
\newblock In F~Pereira, C~J~C Burges, L~Bottou, and K~Q Weinberger, editors,
  \emph{Advances in Neural Information Processing Systems}, volume~25, pages
  1097--1105. Curran Associates, Inc., 2012.
\newblock URL
  \url{https://proceedings.neurips.cc/paper/2012/file/c399862d3b9d6b76c8436e924a68c45b-Paper.pdf}.

\bibitem[Silver et~al.(2016)Silver, Huang, Maddison, Guez, Sifre, van~den
  Driessche, Schrittwieser, Antonoglou, Panneershelvam, Lanctot, Dieleman,
  Grewe, Nham, Kalchbrenner, Sutskever, Lillicrap, Leach, Kavukcuoglu, Graepel,
  and Hassabis]{Silver2016-oh}
David Silver, Aja Huang, Chris~J Maddison, Arthur Guez, Laurent Sifre, George
  van~den Driessche, Julian Schrittwieser, Ioannis Antonoglou, Veda
  Panneershelvam, Marc Lanctot, Sander Dieleman, Dominik Grewe, John Nham, Nal
  Kalchbrenner, Ilya Sutskever, Timothy Lillicrap, Madeleine Leach, Koray
  Kavukcuoglu, Thore Graepel, and Demis Hassabis.
\newblock Mastering the game of go with deep neural networks and tree search.
\newblock \emph{Nature}, 529\penalty0 (7587):\penalty0 484--489, January 2016.
\newblock ISSN 0028-0836.
\newblock \doi{10.1038/nature16961}.
\newblock URL \url{https://www.nature.com/articles/nature16961}.

\bibitem[{OpenAI}(2023)]{OpenAI2023-hh}
{OpenAI}.
\newblock {GPT-4} technical report.
\newblock March 2023.
\newblock URL \url{http://arxiv.org/abs/2303.08774}.

\bibitem[Kingma and Ba(2014)]{Kingma2014-pi}
Diederik~P Kingma and Jimmy Ba.
\newblock Adam: A method for stochastic optimization.
\newblock December 2014.
\newblock URL \url{http://arxiv.org/abs/1412.6980}.

\bibitem[Blei et~al.(2003)Blei, Ng, and Jordan]{blei2003latent}
David~M Blei, Andrew~Y Ng, and Michael~I Jordan.
\newblock Latent dirichlet allocation.
\newblock \emph{Journal of machine Learning research}, 3\penalty0
  (Jan):\penalty0 993--1022, 2003.

\bibitem[Mikolov et~al.(2013)Mikolov, Chen, Corrado, and
  Dean]{mikolov2013efficient}
Tomas Mikolov, Kai Chen, Greg Corrado, and Jeffrey Dean.
\newblock Efficient estimation of word representations in vector space.
\newblock \emph{arXiv preprint arXiv:1301.3781}, 2013.

\bibitem[Vaswani et~al.(2017)Vaswani, Shazeer, Parmar, Uszkoreit, Jones, Gomez,
  Kaiser, and Polosukhin]{vaswani2017attention}
Ashish Vaswani, Noam Shazeer, Niki Parmar, Jakob Uszkoreit, Llion Jones,
  Aidan~N Gomez, {\L}ukasz Kaiser, and Illia Polosukhin.
\newblock Attention is all you need.
\newblock \emph{Advances in neural information processing systems}, 30, 2017.

\bibitem[Wang et~al.(2019{\natexlab{b}})Wang, Li, Xiao, Zhu, Li, Wong, and
  Chao]{wang2019learning}
Qiang Wang, Bei Li, Tong Xiao, Jingbo Zhu, Changliang Li, Derek~F Wong, and
  Lidia~S Chao.
\newblock Learning deep transformer models for machine translation.
\newblock \emph{arXiv preprint arXiv:1906.01787}, 2019{\natexlab{b}}.

\bibitem[Wang et~al.(2019{\natexlab{c}})Wang, Li, and Smola]{wang2019language}
Chenguang Wang, Mu~Li, and Alexander~J Smola.
\newblock Language models with transformers.
\newblock \emph{arXiv preprint arXiv:1904.09408}, 2019{\natexlab{c}}.

\bibitem[Sanh et~al.(2019)Sanh, Debut, Chaumond, and Wolf]{sanh2019distilbert}
Victor Sanh, Lysandre Debut, Julien Chaumond, and Thomas Wolf.
\newblock Distilbert, a distilled version of bert: smaller, faster, cheaper and
  lighter.
\newblock \emph{arXiv preprint arXiv:1910.01108}, 2019.

\bibitem[Breiman(2001)]{Breiman2001-kb}
Leo Breiman.
\newblock Random forests.
\newblock \emph{Mach. Learn.}, 45\penalty0 (1):\penalty0 5--32, October 2001.
\newblock ISSN 0885-6125, 1573-0565.
\newblock \doi{10.1023/A:1010933404324}.
\newblock URL \url{https://doi.org/10.1023/A:1010933404324}.

\bibitem[GRO(2008--2023)]{GROBID}
Grobid.
\newblock \url{https://github.com/kermitt2/grobid}, 2008--2023.

\bibitem[SCI()]{SCIENCEBEAM}
Sciencebeam parser.
\newblock
  \url{https://https://gitlab.coko.foundation/sciencebeam/sciencebeam-parser}.

\bibitem[BEA()]{BEAUTIFULSOUP}
Beautiful soup.
\newblock \url{https://beautiful-soup-4.readthedocs.io/en/latest/}.

\bibitem[Rehurek and Sojka(2011)]{rehurek2011gensim}
Radim Rehurek and Petr Sojka.
\newblock Gensim--python framework for vector space modelling.
\newblock \emph{NLP Centre, Faculty of Informatics, Masaryk University, Brno,
  Czech Republic}, 3\penalty0 (2), 2011.

\bibitem[Wolf et~al.(2020)Wolf, Debut, Sanh, Chaumond, Delangue, Moi, Cistac,
  Rault, Louf, Funtowicz, Davison, Shleifer, von Platen, Ma, Jernite, Plu, Xu,
  Scao, Gugger, Drame, Lhoest, and Rush]{wolf-etal-2020-Transformers}
Thomas Wolf, Lysandre Debut, Victor Sanh, Julien Chaumond, Clement Delangue,
  Anthony Moi, Pierric Cistac, Tim Rault, Rémi Louf, Morgan Funtowicz, Joe
  Davison, Sam Shleifer, Patrick von Platen, Clara Ma, Yacine Jernite, Julien
  Plu, Canwen Xu, Teven~Le Scao, Sylvain Gugger, Mariama Drame, Quentin Lhoest,
  and Alexander~M. Rush.
\newblock Transformers: State-of-the-art natural language processing.
\newblock In \emph{Proceedings of the 2020 Conference on Empirical Methods in
  Natural Language Processing: System Demonstrations}, pages 38--45, Online,
  October 2020. Association for Computational Linguistics.
\newblock URL \url{https://www.aclweb.org/anthology/2020.emnlp-demos.6}.

\bibitem[Gardner and Wackerle(1965)]{gardner1965interactions}
SD~Gardner and Jerry Wackerle.
\newblock Interactions of detonation waves in condensed explosives.
\newblock Technical report, Los Alamos Scientific Lab., Univ. of California, N.
  Mex., 1965.

\bibitem[Wattenberg et~al.(2016)Wattenberg, Viégas, and
  Johnson]{wattenberg2016how}
Martin Wattenberg, Fernanda Viégas, and Ian Johnson.
\newblock How to use t-sne effectively.
\newblock \emph{Distill}, 2016.
\newblock \doi{10.23915/distill.00002}.
\newblock URL \url{http://distill.pub/2016/misread-tsne}.

\bibitem[Devlin et~al.(2018)Devlin, Chang, Lee, and Toutanova]{devlin2018bert}
Jacob Devlin, Ming-Wei Chang, Kenton Lee, and Kristina Toutanova.
\newblock Bert: Pre-training of deep bidirectional transformers for language
  understanding.
\newblock \emph{arXiv preprint arXiv:1810.04805}, 2018.

\bibitem[Singhal et~al.(2023)Singhal, Azizi, Tu, Mahdavi, Wei, Chung, Scales,
  Tanwani, Cole-Lewis, Pfohl, et~al.]{singhal2023large}
Karan Singhal, Shekoofeh Azizi, Tao Tu, S~Sara Mahdavi, Jason Wei, Hyung~Won
  Chung, Nathan Scales, Ajay Tanwani, Heather Cole-Lewis, Stephen Pfohl, et~al.
\newblock Large language models encode clinical knowledge.
\newblock \emph{Nature}, pages 1--9, 2023.

\bibitem[Chowdhery et~al.(2022)Chowdhery, Narang, Devlin, Bosma, Mishra,
  Roberts, Barham, Chung, Sutton, Gehrmann, et~al.]{chowdhery2022palm}
Aakanksha Chowdhery, Sharan Narang, Jacob Devlin, Maarten Bosma, Gaurav Mishra,
  Adam Roberts, Paul Barham, Hyung~Won Chung, Charles Sutton, Sebastian
  Gehrmann, et~al.
\newblock Palm: Scaling language modeling with pathways.
\newblock \emph{arXiv preprint arXiv:2204.02311}, 2022.

\bibitem[Brown et~al.(2020{\natexlab{b}})Brown, Mann, Ryder, Subbiah, Kaplan,
  Dhariwal, Neelakantan, Shyam, Sastry, Askell, et~al.]{brown2020language}
Tom Brown, Benjamin Mann, Nick Ryder, Melanie Subbiah, Jared~D Kaplan, Prafulla
  Dhariwal, Arvind Neelakantan, Pranav Shyam, Girish Sastry, Amanda Askell,
  et~al.
\newblock Language models are few-shot learners.
\newblock \emph{Advances in neural information processing systems},
  33:\penalty0 1877--1901, 2020{\natexlab{b}}.

\end{thebibliography}

\appendix

\clearpage
\section{Model Hyperparameters}
\label{sec:appendix}

\subsection{LDA Model Hyperparameters}

The final, best performing, LDA model (trained with the GENSIM Python library) hyperparameters are provided in Table \ref{tab:lda_hyperparams}.

\begin{table}[H]
	\centering
	\begin{tabular}{p{5cm}p{2cm}}
		Number of Topics & 300 \\[0.5cm]
        Document Topic Prior & 1.0 \\[0.5cm]
        Topic Word Prior & \nicefrac{1}{300} \\[0.5cm]
        Chunksize & 2000 \\[0.5cm]
        Passes & 20 \\[0.5cm]
        Iterations & 400\\
		\bottomrule
	\end{tabular}
    \caption{LDA Model Hyperparameters}
	\label{tab:lda_hyperparams}
\end{table}

\subsection{W2V Model Hyperparameters}

The final, best performing, \gls{w2v} model (trained with the GENSIM Python library) hyperparameters are provided in Table \ref{tab:w2v_hyperparams}

\begin{table}[H]
	\centering
	\begin{tabular}{p{5cm}p{5cm}}
		Embedding Dimension & 300 \\[0.5cm]
        Context Window Size & 2 \\[0.5cm]
        Minimum Word Count & 20 \\[0.5cm]
        Model Variant & Continuous Bag of Words \\
		\bottomrule
	\end{tabular}
    \caption{W2V Model Hyperparameters}
	\label{tab:w2v_hyperparams}
\end{table}

\subsection{Transformer Model Hyperparameters}

The final, best performing, Transformer model (trained with the HuggingFace library) hyperparameters are provided in Table \ref{tab:Transformer_hyperparams}. Hyperparameters not explicitly given are set to their default values according to the Transformer library version 4.24.0.

\begin{table}[H]
	\centering
	\begin{tabular}{p{5cm}p{5cm}}
		Base Transformer Model & distilbert-base-uncased \\[0.5cm]
        Number of Epochs & 100 \\[0.5cm]
        Train Batch Size & 32 \\[0.5cm]
        Tokens per Sample & 512 \\[0.5cm]
        Vocabulary Size & 30522 \\
		\bottomrule
	\end{tabular}
    \caption{Transformer Model Hyperparameters}
	\label{tab:Transformer_hyperparams}
\end{table}

\subsection{Random Forest Model Hyperparameters}

The random forest classification model hyperparameters are provided in Table \ref{tab:rf_hyperparams}.

\begin{table}[H]
	\centering
	\begin{tabular}{p{5cm}p{4cm}}
		Test Size & 0.33 \\[0.5cm]
        Criterion & Gini Coefficient \\[0.5cm]
        Max Features & Sqrt \\[0.5cm]
        Number of Estimators  & 200 \\
		\bottomrule
	\end{tabular}
    \caption{Random Forest Model Hyperparameters}
	\label{tab:rf_hyperparams}
\end{table}

\end{document}